\documentclass{article}

\usepackage{amsmath}
\usepackage{amssymb}
\usepackage{multirow}
\usepackage{graphicx}
\usepackage[most]{tcolorbox}
\usepackage{tcolorbox}
\tcbuselibrary{listings}

\newtcblisting{promptbox}[1][]{
    colback=gray!10,      
    colframe=gray!50,     
    boxrule=0.5mm,        
    arc=1mm,              
    boxsep=0mm,
    width=0.99\textwidth, 
    title=#1,
    fonttitle=\ttfamily\footnotesize\centering,
    listing only,         
    listing options={
        basicstyle=\ttfamily\tiny,
        breaklines=true,
        columns=fullflexible,
        keepspaces=true,
        showstringspaces=false
    }
}
 \usepackage[preprint]{neurips_2026}

\usepackage[utf8]{inputenc} 
\usepackage[T1]{fontenc}    
\usepackage{hyperref}       
\usepackage{url}            
\usepackage{booktabs}       
\usepackage{amsfonts}       
\usepackage{nicefrac}       
\usepackage{microtype}      
\usepackage{xcolor}         
\usepackage{cleveref}

\title{DRT: Dense Reasoning Trace for Efficient and Grounded Multimodal Reasoning}

\author{%
{Wan Xu$^{1}$ ~ Yuanfan Guo{$^{2}$}} ~ \textbf{Kevin Han{$^{3}$} ~ LaLa Chen{$^{4}$} ~ Wangmeng Zuo{$^{1}$\thanks{Corresponding author.}}} \\
\normalsize
\vspace{2pt}
$^{1}$\	Harbin Institute of Technology  
$^{2}$\ Shanghai Jiao Tong University \\
$^{3}$\ Facebook
$^{4}$\ University of California, Irvine
}

\begin{document}

\maketitle
\begin{abstract}
Despite the remarkable progress in Multimodal Large Language Models (MLLMs), prevailing Chain-of-Thought (CoT) paradigms remain largely confined to the space of verbose, free-form natural-language reasoning. Consequently, they inherently incur excessive linguistic overhead, leading to information dilution and weak visual grounding.
To address this challenge, we propose \textbf{Dense Reasoning Trace (DRT)}, a paradigm that departs from verbose, free-form CoT by expressing reasoning as compact structured traces, which include concise intermediate states with symbolic connectors and disentangle visual observations from logical deductions. First, we introduce the \textbf{Dense Trace Initialization} to internalize the DRT reasoning mode into the model, substantially improving token efficiency while preserving visual evidence. To further enable the model to faithfully capture the logical relations within traces, we propose the \textbf{Trace-Grounded Reinforcement Learning} framework, which builds reference traces through a tri-perspective verification pipeline and employs Trace-Grounded GRPO with structured rewards, encouraging the model to generate concise DRT-style traces with reduced hallucination and stronger logical grounding. 
Extensive experiments on challenging reasoning benchmarks show that DRT achieves 5.5$\times$ token efficiency improvement while improving 1.3 accuracy points over the Qwen3-VL baseline. These findings suggest that complex multimodal reasoning may not require verbose natural-language traces, opening a more efficient path for next-generation MLLMs.
\end{abstract}
\section{Introduction}
Chain-of-thought (CoT) reasoning has become a key ingredient in the problem-solving capabilities of Large Language Models (LLMs)~\cite{wei2022chain,guo2025deepseek,bai2025qwen3}. Its success has quickly carried over to multimodal settings, where Multimodal Large Language Models (MLLMs) benefit from explicit intermediate reasoning on tasks such as visual mathematical reasoning~\cite{lu2024mathvista,zhang2024mathverse}, logical deduction~\cite{xiao2024logicvista}, and long-horizon video understanding~\cite{cheng2025video_holmes}. Recent studies further suggest that scaling test-time compute, often through longer reasoning traces and larger token budgets, can substantially improve performance on challenging benchmarks~\cite{feng2023towards,li2024chain,snell2025scaling}. As a result, generating long and detailed reasoning trajectories has become a dominant paradigm in modern multimodal reasoning systems~\cite{OpenAI_o3,anthropic_claude_opus_4_6}.

Despite its effectiveness, CoT reasoning suffers from a fundamental limitation: it relies on verbose, free-form narration, which inherently incurs excessive linguistic overhead. This overhead leads to \emph{informational dilution}, where redundant verbal scaffolding dilutes the density of critical deductions, increasing inference cost while reducing the salience of key reasoning steps~\cite{li2025compressing,she2025hawkeye}. This issue is further exacerbated in multimodal reasoning, where perception and deduction are tightly coupled. As reasoning traces become longer and increasingly self-referential, the model tends to attend more to its generated text rather than visual evidence, amplifying hallucinations and leading to \emph{weak grounding in visual evidence}~\cite{huang2024opera}.

Recent works have recognized the substantial computational cost of CoT inference and have mainly focused on reducing output token usage. Training-free approaches improve efficiency via salient token selection~\cite{zhang2026chain}, early stopping~\cite{li2025thinkless}, or concise prompting~\cite{xu2025cod}, while training-based methods learn compact reasoning through adaptive token skipping~\cite{xia2025tokenskip} or token-budget estimation~\cite{li2025selfbudgeter}. However, these methods still rely on free-form textual reasoning, leaving the underlying source of inefficiency largely unaddressed and limiting the accuracy-efficiency tradeoff.

\begin{figure}[t]
    \centering
    \includegraphics[width=0.87\linewidth]{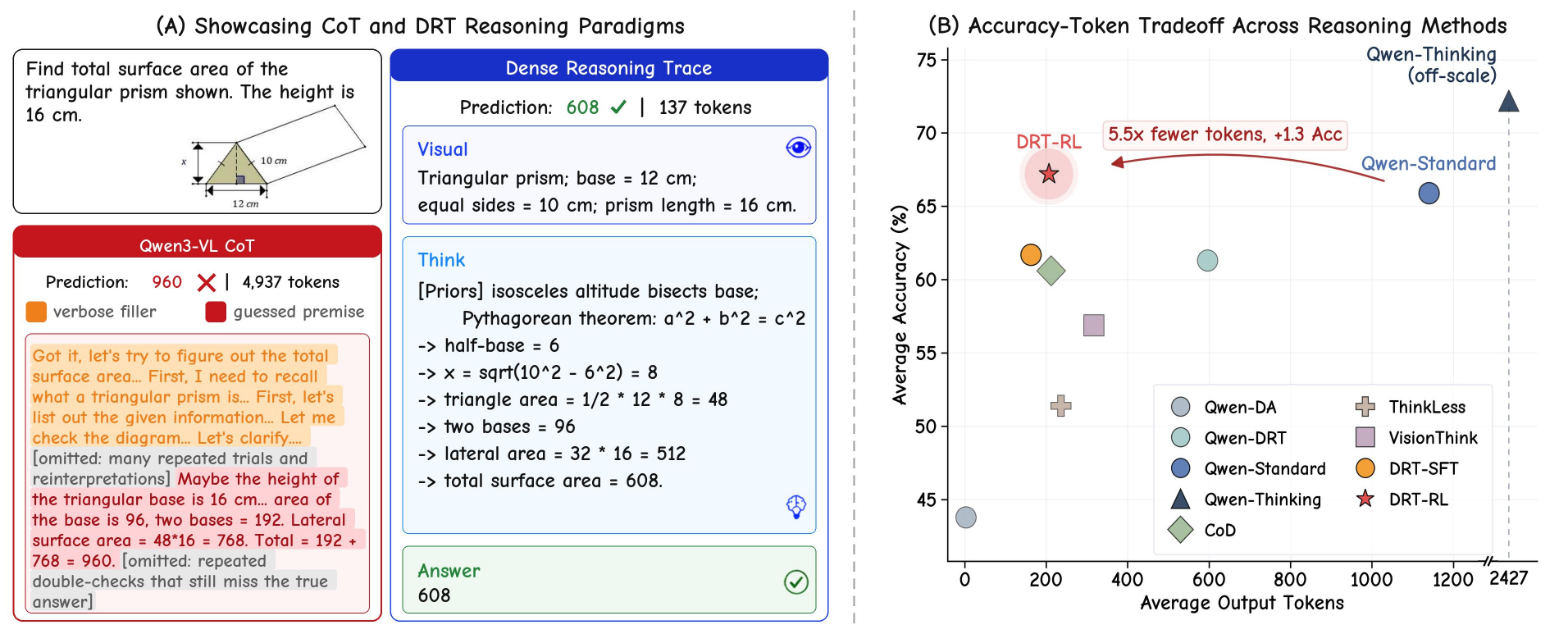}
    \vspace{-4mm}
    \caption{\textbf{Qualitative and quantitative comparison of CoT and DRT reasoning paradigms}. \textbf{(A)}: Qwen3-VL produces an overly long, low-density trace with an image-unfaithful assumption, leading to an incorrect answer. In contrast, DRT-RL successfully solves the problem using structured, compact reasoning traces. \textbf{(B)}: Across prompting paradigms and efficient-reasoning baselines, DRT-RL achieves a stronger accuracy-token tradeoff.}
    \label{fig:teaser}
    \vspace{-6mm}
\end{figure}

To address this challenge, we propose \textbf{Dense Reasoning Trace (DRT)}, a paradigm that departs from verbose, free-form CoT, as shown in Figure~\ref{fig:teaser}(A). DRT represents reasoning as concise intermediate states with symbolic connectors, reducing inference cost while preserving the salience of key conclusions and logical relations. It further explicitly disentangles visual observations from logical deduction, yielding a structured reasoning process that prevents assumptions from being conflated with observed evidence. In this way, DRT directly addresses the limitation of CoT by reorganizing explicit reasoning, mitigating informational dilution and improving visual grounding.

We first introduce \textbf{Dense Trace Initialization}, which internalizes the DRT reasoning mode into the model, substantially improving token efficiency while preserving visual evidence. In this stage, we first construct the \textbf{DRT-SFT} dataset by rewriting high-quality multimodal instruction data into the dense trace format. In this format, visual evidence is organized in \texttt{<visual>}, while compact, prior-guided logical derivations are represented in \texttt{<think>}. This supervised stage enables the model to internalize the DRT reasoning mode and serves as initialization for subsequent training.

To further enable the model to faithfully capture the logical relations within traces, we propose the \textbf{Trace-Grounded Reinforcement Learning} framework, improving reasoning faithfulness under tight token budgets. To support this stage, we construct a \textbf{DRT-RL} dataset through a tri-perspective verification pipeline, where only traces with correct final answers, logically consistent intermediate steps, and faithful visual grounding are retained. Using these verified traces as reference, we optimize the model with Trace-Grounded Group Relative Policy Optimization (GRPO), together with process-based rewards and exploratory incentives. This stage encourages the model to search for accurate deductive paths while remaining concise and grounded.

Extensive experiments validate the effectiveness of our approach. Across five challenging reasoning benchmarks, including text and out-of-distribution video tasks, our DRT-based model achieves 5.5$\times$ token efficiency improvement while improving 1.3 accuracy points over Qwen3-VL-8B-Instruct under official prompting, as shown in Figure~\ref{fig:teaser}(B). These results demonstrate the robustness of DRT across diverse multimodal reasoning settings. Comprehensive ablation studies further confirm the necessity of each core component. 
In summary, our main contributions are as follows:
\begin{list}{\labelitemi}{\leftmargin=10pt}
\item We propose DRT, a multimodal reasoning paradigm that replaces verbose, free-form CoT with compact structured traces to reduce linguistic overhead. By reformulating the surface organization of explicit reasoning, DRT mitigates informational dilution and improves visual grounding.
\item We develop a two-stage training framework to implement the DRT paradigm, including Dense Trace Initialization, which enables the model to internalize the DRT reasoning mode, and Trace-Grounded Reinforcement Learning, which enables the model to faithfully capture the logical relations within traces, yielding efficient and accurate reasoning under tight token budgets.
\item Our experiments show that DRT achieves a 5.5$\times$ reduction in token usage while improving accuracy by 1.3 points over the Qwen3-VL baseline, yielding a superior efficiency-accuracy tradeoff and suggesting a more efficient path for next-generation MLLMs.
\end{list}
\vspace{-2.5mm}
\section{Related Work}
\noindent\textbf{Reasoning Models and Test-Time Scaling.}
Chain-of-Thought prompting~\cite{wei2022chain} has substantially improved the reasoning ability of LLMs. Building on this paradigm, prior work has explored various reasoning frameworks, including planning-based methods~\cite{yao2023tree,besta2024graph} and search-based decoding strategies~\cite{chen2024alphamath}. More recently, DeepSeek-R1~\cite{guo2025deepseek} showed that large-scale RL, combined with format and outcome rewards, can induce strong reasoning behavior. In parallel, recent theoretical and empirical studies have formalized test-time scaling, showing that additional inference-time compute can yield substantial gains beyond parameter scaling alone~\cite{feng2023towards,li2024chain,snell2025scaling}. These advances have motivated the development of Multimodal Large Language Models (MLLMs)~\cite{huang2025vision,OpenAI_o3} for challenging reasoning-intensive multimodal tasks~\cite{zhang2024mathverse,xiao2024logicvista}. However, longer reasoning traces also introduce clear drawbacks, including higher computational cost and increased exposure to hallucination over extended trajectories.

\noindent\textbf{Efficient Thinking and Reasoning.}
Recent work improves reasoning efficiency along two complementary directions. One line performs intermediate computation in continuous states through recursive hidden-state feedback~\cite{hao2025training}, explicit-to-latent self-distillation~\cite{shen2025codi}, or projected soft thoughts~\cite{xu2025softcot}; extensions introduce an auxiliary decoder for step-level supervision~\cite{wei2025sim} or vision-aligned latent tokens~\cite{jeon2026vision}. These methods reduce explicit token decoding, but their intermediate states are not natively exposed as human-readable steps, and semantic process supervision often requires additional alignment or decoding mechanisms.
A complementary line retains explicit textual reasoning while reducing inference cost through visually salient token selection~\cite{zhang2026chain}, early stopping~\cite{li2025thinkless}, and concise prompting~\cite{xu2025cod}, or through training with adaptive token skipping~\cite{xia2025tokenskip}, token-budget estimation~\cite{li2025selfbudgeter}, and reasoning path selection~\cite{huang2026learning}.
DRT remains an explicit textual representation, but replaces free-form narration with compact state transitions and separates visual evidence from logical deduction. This design reduces narrative overhead while preserving direct process inspection and step-level supervision.
\section{Method}\label{method}
\begin{figure}[t]
    \centering
    \includegraphics[width=0.95\linewidth]{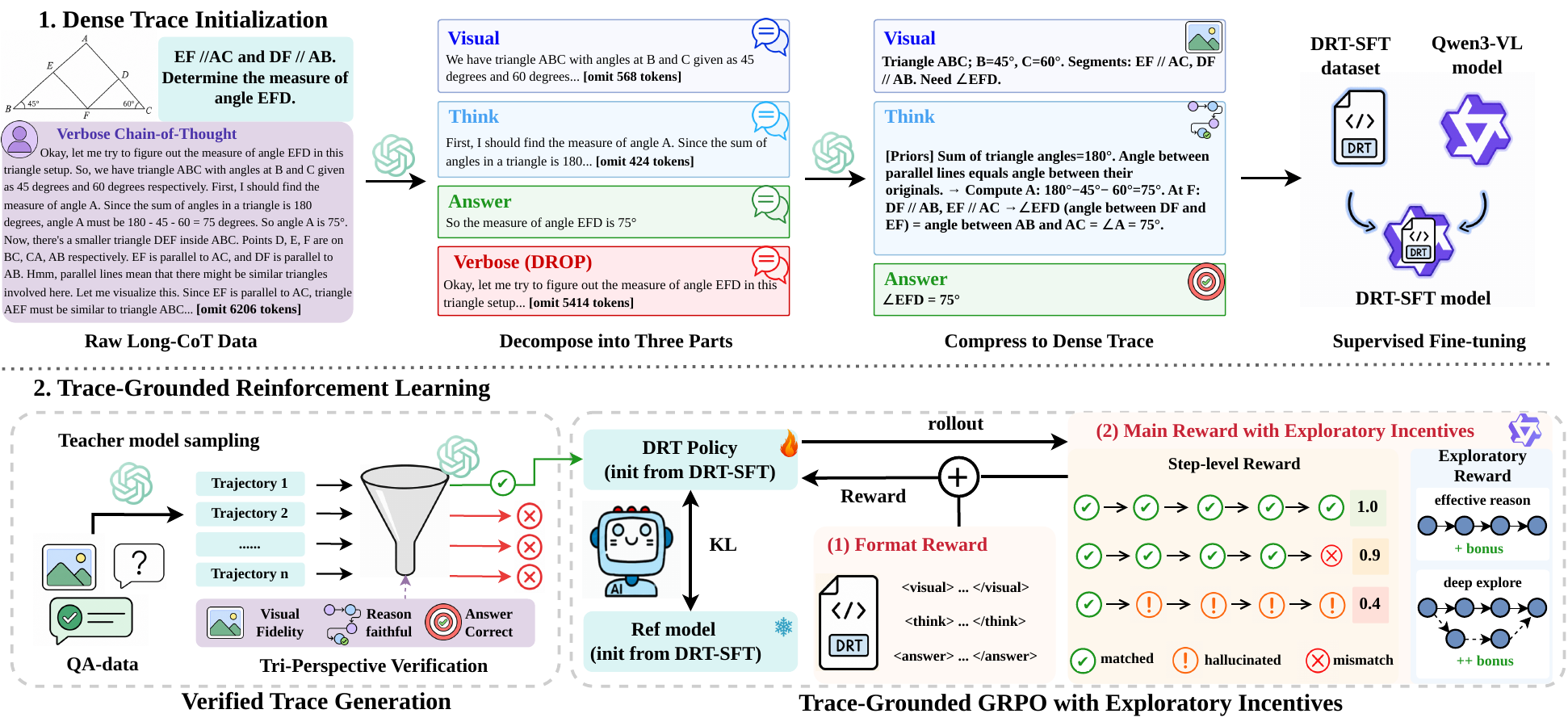}
    \vspace{-3mm}
    \caption{The overall pipeline of DRT.}
    \label{fig:method}
    \vspace{-6mm}
\end{figure}

In this section, we present \textbf{Dense Reasoning Trace (DRT)}, a paradigm for improving multimodal reasoning efficiency by compressing verbose reasoning trajectories into compact, high-density structured traces. The overall pipeline is shown in Figure~\ref{fig:method}. We first formalize the DRT representation in Sec.~\ref{drt_define}, where visual perception is explicitly separated from logical deduction and the latter is organized into a concise format. We then introduce a two-stage optimization pipeline: 
(\textit{i}) \textbf{Dense Trace Initialization} (Sec.~\ref{sft}), where raw instruction data are reformulated into the dense trace format to initialize the model with efficient reasoning patterns.
(\textit{ii}) \textbf{Trace-Grounded Reinforcement Learning} (Sec.~\ref{rl}), where a tri-perspective verification pipeline is used to construct the DRT-RL dataset with grounded stepwise traces, followed by Trace-Grounded GRPO with process-based rewards and exploratory incentives to improve reasoning accuracy under tight token budgets.

\subsection{Dense Reasoning Trace Representation}\label{drt_define}
Although long CoT reasoning has shown strong performance on complex multimodal tasks, it is fundamentally constrained by its reliance on verbose natural-language narration, which incurs substantial linguistic overhead and gives rise to two key issues: \emph{informational dilution} and \emph{weakened visual grounding}.
To maintain fluency, models often generate substantial verbal scaffolding that contributes little to the actual reasoning process, resulting in unnecessary computational overhead~\cite{li2025compressing,she2025hawkeye} and even degenerate repetitive loops~\cite{NEURIPS2023_e6c2e85d,yao2025understanding}. In multimodal settings, long and increasingly self-referential reasoning traces may also cause the model to rely more on its own generated text than on the original visual evidence, thereby weakening grounding fidelity and increasing the risk of hallucination~\cite{huang2024opera}.

To address these limitations, we propose \textbf{Dense Reasoning Trace (DRT)}, a reasoning paradigm with a structured representation designed to improve both reasoning efficiency and perceptual faithfulness. Unlike standard CoT, which emphasizes natural-language fluency, DRT prioritizes information density and explicit grounding. Specifically, DRT compresses the reasoning trajectory into a high-density format using symbolic connectors, replacing verbose narration with compact reasoning and corresponding intermediate conclusions. To preserve grounding under this compressed representation, DRT further disentangles visual perception from logical deduction, so that observed evidence and inferred conclusions are represented in distinct stages. As a result, each generated token is encouraged to serve as either a grounded observation or a high-value reasoning step. Accordingly, we organize the reasoning process into three structured components:

\noindent\textbf{Visual Grounding (\texttt{<visual>}):}
This component aggregates essential visual primitives, including object attributes, spatial relations, counts, and temporal cues, into an explicit evidence set. By isolating perceptual evidence from downstream inference, it provides a stable grounding anchor and reduces the risk of later deductions drifting away from the visual input.

\noindent\textbf{Concise Deduction (\texttt{<think>}):}
Given the grounded visual evidence, this component first introduces task-relevant priors through an explicit \texttt{[Priors]} token, such as mathematical formulas, commonsense knowledge, and predefined rules. Deduction is then carried out jointly over the visual evidence and these priors as a compressed sequence of high-information state transitions. Rather than generating fluent but redundant explanations, DRT represents the reasoning process in a concise symbolic format, retaining only essential relations, constraints, and intermediate conclusions. This design improves the information-to-token ratio and reduces computational overhead.

\noindent\textbf{Conclusion (\texttt{<answer>}):}
The final answer gives a concise response to the original question, summarizing
the result derived from the visual observations and logical reasoning.


Overall, DRT reformulates multimodal reasoning as a sequence of discrete, high-information states with explicit separation between perception and deduction. This representation not only improves reasoning efficiency by reducing linguistic redundancy, but also enhances faithfulness by encouraging deductions to remain anchored to verifiable visual evidence.

\subsection{Dense Trace Initialization}\label{sft}
Supervised fine-tuning (SFT) has been shown to be effective in improving instruction following~\cite{ouyang2022training,wang2023self} and shaping the reasoning behavior of large models~\cite{huang2025vision,xia2025tokenskip,jiang2025drp}. Motivated by this, we construct a dedicated DRT-SFT dataset $\mathcal{D}_{\text{SFT}}$ by reformulating 200K multimodal samples from the Mulberry subset of the Vision-R1-Cold dataset~\cite{huang2025vision}, leveraging its high-quality reasoning trajectories across diverse multimodal tasks. We then apply a \textit{Deconstruct-and-Compress} pipeline to transform the original responses into the DRT format. Specifically, we use GPT-5.1~\cite{singh2025openai} as the teacher model to perform a three-step transformation: (\textit{i}) \textbf{Deconstructive Separation:} each original response is decomposed into three mutually exclusive components: \textit{Visual Observations}, \textit{Logical Reasoning}, and the \textit{Final Answer}. (\textit{ii}) \textbf{Concise Compression:} natural-language narration is rewritten into a dense reasoning trace by replacing linguistic fillers with symbolic connectors (e.g., $\rightarrow$) and concise phrases. (\textit{iii}) \textbf{Explicit Anchoring:} within the $\texttt{<think>}$ module, we enforce a structured progression beginning with a $\texttt{[Priors]}$ anchor. This anchor encapsulates domain-specific formulas, commonsense axioms, or geometric constraints (e.g., $Area = \pi r^2$) distilled from the original trajectory and task context. Together with the extracted $\texttt{<visual>}$ evidence, it forms a complete set of preliminary information that provides the epistemic grounding for subsequent logical transitions. The detailed DRT-SFT dataset generation prompt is provided in Appendix~\ref{app:sft_dataset_prompt}.



\subsection{Trace-Grounded Reinforcement Learning}\label{rl}
Although SFT provides a stable initialization and equips the model with the basic DRT reasoning format, overly short or aggressively pruned reasoning traces may remove critical information or disrupt the logical chain~\cite{nohara2026optimal,xia2025tokenskip}. This issue is amplified by the stricter step-level requirements of DRT, where minor early errors can propagate through the compact trace.

To mitigate such error accumulation and improve the robustness of intermediate reasoning, we introduce a reinforcement learning (RL) stage with step-level supervision, which has been shown to help correct errors in intermediate reasoning steps~\cite{lightman2023let,setlur2025rewarding}. This paradigm aligns naturally with DRT, whose structured representation decomposes reasoning into atomic steps, enabling precise step-level supervision. Building on this, we propose the Trace-Grounded Reinforcement Learning framework to enhance reasoning reliability while preserving the compactness of DRT. Specifically, we construct a DRT-RL dataset $\mathcal{D}_{\text{RL}}$ consisting of verified stepwise trajectories, and optimize the SFT-initialized model using our designed Trace-Grounded GRPO with exploratory incentives. 

\subsubsection{Verified Trace Generation}\label{RL-dataset}
Existing multimodal RL datasets predominantly provide only final answers without reliable stepwise annotations~\cite{huang2025vision,yu2025dapo}, making them incompatible with the construction paradigm of DRT-SFT, which relies on detailed long-form reasoning traces. To address this limitation, we construct the RL dataset $\mathcal{D}_{\text{RL}}$ via a tri-perspective verification pipeline that augments answer-only data with verified stepwise reasoning trajectories using a strong teacher model and a dedicated verification mechanism.

Formally, let $(q, a^*) \sim \mathcal{D}_{\text{raw}}$ denote a question-answer pair, where $q$ may include textual input and an associated image, and $a^*$ is the ground-truth answer. For each $q$, we use a strong teacher model (GPT-5.1~\cite{singh2025openai}) to sample a set of candidate reasoning trajectories $\mathcal{S}(q) = \{ S^{(1)}, \dots, S^{(K)} \}$. Each trajectory $S^{(k)} = (s^{(k)}_1, \dots, s^{(k)}_{T_k})$ is a sequence of atomic reasoning units, where each $s^{(k)}_t$ represents a logical transition and its intermediate conclusion, enabling fine-grained supervision. 

To retain reliable and informative optimization targets, we further employ GPT-5.1 as a strict verifier $\mathcal{V}(\cdot)$ to evaluate each trajectory $S^{(k)}$ conditioned on $(q, a^*)$. In practice, final-answer correctness alone is insufficient to ensure reasoning quality. A trajectory may contain \textit{visual hallucinations}, where unverified quantities inferred from the image are treated as implicit premises, or \textit{reasoning hallucinations}, where unsupported conditions, relations, or intermediate values are introduced to bridge gaps in deduction. Although such trajectories may occasionally reach the correct final answer, their hallucinated or unsupported premises make them unreliable for step-level supervision in subsequent RL training. To address these issues, the verifier first checks the semantic correctness of the final answer and then audits each trajectory along two dimensions:

\noindent\textbf{Visual Fidelity:}
Every perceptual claim must be grounded in explicitly observable evidence from the input, such as annotated text, symbols, or clearly specified visual attributes. This criterion prevents unverified visual estimates or hallucinated observations from being used as implicit premises.

\noindent\textbf{Reasoning Faithfulness:}
Each deductive step must be justified by verified observations, the problem statement, introduced priors, or established theorems. This criterion rules out speculative shortcuts, ad hoc assumptions, circular reasoning, and unsupported intermediate relations.

The verifier outputs a binary decision $\mathcal{V}(q, S^{(k)}, a^*) \in \{\texttt{PASS}, \texttt{FAIL}\}$, and only trajectories that receive a \texttt{PASS} verdict are retained. Each accepted trajectory $S^*$ is paired with its corresponding $(q, a^*)$, forming the RL dataset $\mathcal{D}_{\text{RL}} = \{ (q, S^*, a^*) \mid (q, a^*) \sim \mathcal{D}_{\text{raw}},\; S^* \in \mathcal{S}(q),\; \mathcal{V}(q, S^*, a^*) = \texttt{PASS} \}$. This construction ensures that the retained trajectories are reliable and faithful across perception, intermediate reasoning, and final answer stages, yielding high-fidelity step-level supervision signals for RL training. Detailed DRT-RL dataset generation prompts are provided in Appendix~\ref{app:rl_dataset_prompt}.

Following this pipeline, we construct $\mathcal{D}_{\text{RL}}$ from both multimodal and text-only sources. Specifically, Vision-R1-RL~\cite{huang2025vision} provides 8,936 diverse image-text reasoning samples, while DAPO~\cite{yu2025dapo} contributes 5,788 challenging long-form text-only samples.

\subsubsection{Trace-Grounded GRPO}\label{reward}
To align DRT training with accurate, faithful, and efficient reasoning, we propose Trace-Grounded GRPO, which leverages reference trajectories $S^*$ and a large-scale reward judge to supervise generated trajectory $S$ via a structured reward with step-level supervision and exploration incentives:
\[
R = R_{\text{format}} + R_{\text{main}} + R_{\text{bonus}}
\]
where $R_{\text{format}}$ enforces structured and high-density reasoning, $R_{\text{main}}$ captures answer correctness with partial credit, and $R_{\text{bonus}}$ encourages effective multi-step reasoning and exploration.

To compute the reward, we extract step-level signals by comparing the generated trajectory $S$ with the reference trajectory $S^*$. Specifically, $N_{\text{match}}$ counts reference steps whose core logical transition or
intermediate conclusion is correctly reproduced in $S$. Among the matched candidates, $N_{\text{hall}}$ counts steps whose agreement with the reference is achieved through unsupported assumptions, invented values, or ungrounded visual or logical claims. Beyond reference matching, we also evaluate the intrinsic utility of generated steps. $N_{\text{eff}}$ counts distinct and non-trivial steps in $S$ that are necessary or directly useful for deriving the final solution, even when their granularity or decomposition differs from the reference trajectory used for $N_{\text{match}}$. In contrast, $N_{\text{deep}}$ counts valid and relevant exploratory steps that go beyond the minimal reference solution, such as intermediate verification, elimination of alternatives, or alternative derivations. These statistics are obtained using a strong LLM, Qwen3-235B-A22B-Instruct. The detailed judge prompt is provided in Appendix~\ref{app:verl_judge_prompt}.

\noindent\textbf{Format reward.}
We introduce a format reward $R_{\text{format}}$ to enforce structural validity and high-density reasoning, acting as a structural regularizer that keeps the policy within the DRT representation and prevents degeneration into verbose chain-of-thought reasoning. Specifically, valid outputs must maintain the required structure (e.g., well-formed \texttt{<think>} and \texttt{<answer>} segments) and follow the symbolic, concise style (e.g., using ->). We also penalize overly verbose or poorly segmented traces, encouraging compact and information-dense reasoning.

\noindent\textbf{Main reward.}
The main reward combines outcome-level correctness with step-level partial credit. Process completeness is measured by the proportion of correctly matched reasoning steps, defined as $\rho_{\text{comp}} = N_{\text{match}} / |S^*|$, where $|S^*|$ is the number of reference steps within the reference trajectory $S^*$. Let $a$ denote the answer extracted from the generated trajectory $S$. The main reward is:
\[
R_{\text{main}} = \lambda_{\text{main}} \cdot \mathbb{I}[a = a^{*}] + \lambda_{\text{proc}} \cdot \rho_{\text{comp}} \cdot (1 - \mathbb{I}[a = a^{*}])
\]
where $\mathbb{I}[a = a^{*}]$ is computed using both rule-based matching and LLM-based judgment.

To account for the fact that matched steps may still contain hallucinated derivations, we further apply a multiplicative hallucination correction. Let $\rho_{\text{hall}} = N_{\text{hall}} / \max(N_{\text{match}}, 1)$ denote the fraction of hallucinated matched steps. The corrected reward becomes:
\[
R_{\text{main}} \leftarrow R_{\text{main}} \cdot (1 - \lambda_{\text{hall}}\rho_{\text{hall}})
\]

\noindent\textbf{Step bonus reward.}
After supervised fine-tuning under compact reasoning formats, the model tends to produce homogeneous reasoning trajectories across rollouts, resulting in low reward variance and weak optimization signals. To mitigate this, we introduce a step bonus that encourages trajectories to achieve sufficient reasoning coverage relative to the reference solution while allowing controlled exploration beyond the minimal solution path. Formally, we define:
\[
\rho_{\text{reason}} = \min\!\left(N_{\text{eff}} / \max(|S^*|, 1),\, 1\right) \quad
\rho_{\text{explore}} = \min\!\left(N_{\text{deep}} / N_{\text{cap}},\, 1\right)
\]
where $N_{\text{cap}}$ is a predefined hyperparameter that limits excessive exploration. To further account for problem difficulty, we introduce a scaling factor $\rho_{\text{diff}}$ based on $|S^*|$, encouraging deeper reasoning on more complex instances. The resulting bonus is:
\[
R_{\text{bonus}} =
\lambda_{\text{step}} \cdot \rho_{\text{comp}} \cdot \rho_{\text{reason}}
+
\lambda_{\text{explore}} \cdot \rho_{\text{comp}} \cdot \rho_{\text{diff}} \cdot \rho_{\text{explore}}
\]
This design encourages trajectories that match the reference reasoning steps while permitting limited, difficulty-aware exploration. The inclusion of $\rho_{\text{comp}}$ ensures that such exploration remains grounded in correct intermediate reasoning.

Overall, this reward design combines structural regularization, hallucination-aware process credit, and difficulty-aware step incentives into a unified objective, allowing RL to improve reasoning reliability while preserving compact traces. Detailed hyperparameters are provided in Appendix~\ref{app:verl_hyperparameters}.


\section{Experiments}
\subsection{Experimental Setting}
\noindent\textbf{Training Details.}
During SFT, we fine-tune Qwen3-VL-8B-Instruct~\cite{bai2025qwen3} with a peak learning rate of $6.0 \times 10^{-6}$ and cosine decay to $1.0 \times 10^{-6}$. We use dynamic batching and train for up to 2{,}000 iterations; early stopping is triggered at iteration 800, after processing 369{,}455 training samples. During RL, we optimize the SFT-initialized model with a learning rate of $1.0 \times 10^{-6}$ and a batch size of 512 for 200 steps. Additional training and implementation details are in Appendix~\ref{app:training}.

\noindent\textbf{Compared Methods.}
We compare the proposed DRT framework with a diverse set of baselines spanning conventional prompting strategies and recent efficient reasoning approaches.
For prompting-based baselines, we evaluate Qwen3-VL-8B-Instruct~\cite{bai2025qwen3} under different prompting strategies, \textbf{Qwen-DA} performs direct answering without explicit reasoning; \textbf{Qwen-Standard} follows the default prompting format; \textbf{Qwen-DRT} applies our DRT prompting scheme without training to assess prompting-only effects. We also include \textbf{Qwen-Thinking} as an approximate upper bound of reasoning performance within the Qwen3-VL series. Prompts are provided in Appendix~\ref{app:evaluation_prompts}.
For efficient reasoning methods, we include \textbf{CoD}~\cite{xu2025cod}, which constrains each step to minimal draft-style reasoning, and \textbf{ThinkLess}~\cite{li2025thinkless}, which applies early stopping via a \texttt{</think>} token. In our implementation, we set the maximum output token budget of ThinkLess to 300, which aligns with the typical output lengths of efficient reasoning methods. We also include \textbf{VisionThink}~\cite{yang2025visionthink}, which improves efficiency by adaptively selecting visual input resolution. CoD and ThinkLess are implemented on Qwen3-VL-8B-Instruct, while VisionThink is evaluated using its official checkpoint.

\noindent\textbf{Benchmarks \& Metrics.}
We evaluate DRT on five reasoning benchmarks. MathVista~\cite{lu2024mathvista} and MathVerse~\cite{zhang2024mathverse} assess multimodal mathematical reasoning, while LogicVista~\cite{xiao2024logicvista} targets logical reasoning in visual contexts. We further include Video-Holmes~\cite{cheng2025video_holmes} and GSM8K~\cite{cobbe2021training} to examine whether DRT preserves video and text-only reasoning abilities.
We report both accuracy and efficiency, where efficiency is evaluated using the number of output tokens, latency (seconds per request), and throughput (QPS). More detailed efficiency evaluation settings are provided in Appendix~\ref{app:efficient_eval_setting}.

\begin{table}[t]
\centering
\caption{Performance and token efficiency comparison on five reasoning benchmarks. For each benchmark, we report accuracy (Acc$\uparrow$) and average output tokens (Tokens$\downarrow$). The final columns report the macro average of Acc, Tokens, throughput (QPS$\uparrow$) and Latency (Lat.$\downarrow$).}
\resizebox{1.0\linewidth}{!}{
\begin{tabular}{lcccccccccccccc}
\toprule
\multirow{2}{*}{\textbf{Method}}
& \multicolumn{2}{c}{\textbf{MathVista}}
& \multicolumn{2}{c}{\textbf{MathVerse}}
& \multicolumn{2}{c}{\textbf{LogicVista}}
& \multicolumn{2}{c}{\textbf{GSM8K}}
& \multicolumn{2}{c}{\textbf{Video-Holmes}}
& \multicolumn{4}{c}{\textbf{AVG.}} \\
\cmidrule(lr){2-3} \cmidrule(lr){4-5} \cmidrule(lr){6-7} \cmidrule(lr){8-9} \cmidrule(lr){10-11} \cmidrule(lr){12-15}
& Acc$\uparrow$ & Tokens$\downarrow$
& Acc$\uparrow$ & Tokens$\downarrow$
& Acc$\uparrow$ & Tokens$\downarrow$
& Acc$\uparrow$ & Tokens$\downarrow$
& Acc$\uparrow$ & Tokens$\downarrow$
& Acc$\uparrow$ & Tokens$\downarrow$ 
& QPS$\uparrow$ & Lat.$\downarrow$\\
\midrule

Qwen-DA
& 66.5 & 1.67
& 42.5 & 3.3
& 41.6 & 1.2
& 25.6 & 3.7
& 42.8 & 3.9
& 43.8  & 2.8
& 20.04 & 13.83 \\

Qwen-DRT
& 74.9 & 555.6
& 60.5 & 974.1
& 52.4 & 1124.6
& 75.2 & 124.9
& 43.4 & 200.9
& 61.3 & 596.0
& 1.84 & 29.78 \\

Qwen-Standard
& 77.2 & 820.0
& 62.1 & 1079.7
& 55.3 & 3165.1
& 95.2 & 339.6
& 39.6 & 294.4
& 65.9 & 1139.8
& 1.49 & 57.34 \\

Qwen-Thinking
& 79.9 & 1848.4
& 74.2 & 2665.7
& 68.2 & 4528.6
& 96.1 & 1225.1
& 42.5 & 1868.3
& 72.2 & 2427.2
& 0.69 & 106.57 \\

CoD
& 72.0 & 215.4
& 57.6 & 504.9
& 50.3 & 164.5
& 80.7 & 71.3
& 42.2 & 102.8
& 60.6 & 211.8
& 2.47 & 15.55 \\

ThinkLess
& 66.4 & 238.2
& 39.4 & 242.4
& 29.5 & 292.1
& 82.9 & 195.3
& 38.6 & 211.6
& 51.4 & 235.9
& 5.84 & 16.23 \\

VisionThink
& 67.9 & 253.4
& 49.7 & 415.0
& 45.4 & 306.0
& 85.8 & 274.1
& 35.4   & 337.1
& 56.9   & 317.1
& 1.04 & 65.08 \\

DRT-SFT
& 71.5 & 156.1
& 56.2 & 205.7
& 47.2 & 179.8
& 91.0 & 139.4
& 42.8   & 131.8
& 61.7   & 162.6
& 8.03 & 10.56  \\

DRT-RL
& 76.8 & 198.7
& 64.3 & 275.2
& 56.8 & 229.8
& 94.3 & 175.9
& 43.7   & 155.0
& 67.2   & 206.9 
& 6.91 & 11.62  \\
\bottomrule
\end{tabular}
}
\label{tab:main_results}
\vspace{-6mm}
\end{table}

\subsection{Main Results}
Table~\ref{tab:main_results} summarizes the performance of different methods across five reasoning benchmarks, reporting reasoning accuracy, token efficiency, as well as average QPS and latency. Zero-shot DRT prompting on stronger models, detailed comparisons of efficiency metrics and case-level analysis are provided in Appendix~\ref{app:zero_shot},~\ref{app:efficient_eval_analysis}, and~\ref{app:case_study}.

We first analyze the reasoning behavior of the Qwen model family.
\textbf{Qwen-DA} is highly efficient but suffers a substantial accuracy drop, showing that answer-only prediction is insufficient for complex multimodal reasoning.
\textbf{Qwen-Standard} achieves strong accuracy as a baseline, but requires significantly more output tokens, indicating poor token efficiency of standard chain-of-thought reasoning.
\textbf{Qwen-DRT} improves the accuracy-efficiency trade-off, but still uses $3$--$4\times$ more tokens than DRT-SFT, suggesting that prompting alone is insufficient to induce concise structured reasoning.

We then compare DRT with existing efficient reasoning methods.
\textbf{CoD} reduces average token usage by enforcing minimal draft-style reasoning, but its accuracy remains weaker than DRT-based methods.
\textbf{ThinkLess} further constrains the reasoning budget and improves efficiency, but the strict budget also limits reasoning capacity, leading to degraded accuracy on more challenging benchmarks such as MathVerse.
\textbf{VisionThink} reduces input-side computation through compressed visual inputs, but its overall performance remains weaker than DRT-based methods, suggesting that output-side reasoning compression is a more effective direction for improving multimodal reasoning efficiency.

In contrast, our DRT framework achieves a stronger balance between efficiency and reasoning quality.
\textbf{DRT-SFT} substantially reduces output length while preserving strong overall accuracy, eliminating a large portion of redundant token generation and KV cache accumulation that typically burden autoregressive inference.
\textbf{DRT-RL} further improves accuracy by leveraging high-quality process-supervised data and teacher-guided verification under the constrained DRT format. Notably, it outperforms the strong Qwen-Standard baseline on average while reducing output tokens by more than 80\%, demonstrating that compact reasoning can remain accurate when properly optimized.

\begin{table*}[t]
\centering
\small
\setlength{\tabcolsep}{6pt}
\renewcommand{\arraystretch}{1.15}
\caption{Ablation study on SFT init, RL data composition, and reward design. We report accuracy (Acc$\uparrow$) on five benchmarks and their macro average, along with average output tokens (Tokens$\downarrow$). $R_{\text{ans}}$ denotes the standard GRPO reward based on answer-level supervision.}
\label{tab:ablation_models}
\resizebox{1.0\linewidth}{!}{
\begin{tabular}{lllccccccc}
\toprule
\multirow{2}{*}{\textbf{SFT Init}} 
& \multirow{2}{*}{\textbf{RL Data Composition}} 
& \multirow{2}{*}{\textbf{Reward}}
& \textbf{\textbf{MathVista}}
& \textbf{\textbf{MathVerse}}
& \textbf{\textbf{LogicVista}}
& \textbf{\textbf{GSM8K}}
& \textbf{\textbf{Video-Holmes}}
& \multicolumn{2}{c}{\textbf{AVG.}} \\
\cmidrule(lr){4-4} \cmidrule(lr){5-5} \cmidrule(lr){6-6} \cmidrule(lr){7-7} \cmidrule(lr){8-8} \cmidrule(lr){9-10}
& & & Acc$\uparrow$ & Acc$\uparrow$ & Acc$\uparrow$ & Acc$\uparrow$ & Acc$\uparrow$ & Acc$\uparrow$ & Tokens$\downarrow$ \\
\midrule
VisionR1-cold     & Image+Text             & $R_{\text{ans}}$           & 72.0 & 64.4 & 57.3 & 95.1 & 38.6 & 65.5 & 559.3 \\
VisionR1-cold     & Image+Text             & $R_{\text{format}} + R_{\text{main}} + R_{\text{bonus}}$    & 72.3 & 63.4 & 57.7 & 95.5 & 41.4 & 66.1 & 626.3 \\
VisionR1-dense     & Image+Text             & $R_{\text{main}} + R_{\text{bonus}}$    & 73.8 & 58.5 & 53.0 & 94.8 & 42.6 & 64.5 & 253.5 \\
DRT-SFT           & Image-only             & $R_{\text{format}} + R_{\text{main}} + R_{\text{bonus}}$    & 75.5 & 64.2 & 53.7 & 93.6 & 43.1 & 66.0 & 212.2 \\
DRT-SFT           & Image+Text             & $R_{\text{format}} + R_{\text{ans}}$           & 74.4 & 62.0 & 50.8 & 92.7 & 42.5 & 64.5 & 159.8 \\
DRT-SFT           & Image+Text             & $R_{\text{format}} + R_{\text{main}}$  & 74.5 & 63.1 & 54.1 & 94.1 & 44.8 & 66.1 & 170.4 \\
DRT-SFT           & Image+Text             & $R_{\text{format}} + R_{\text{main}} + R_{\text{bonus}}$    & 76.8 & 64.3 & 56.8 & 94.3 & 43.7 & 67.2 & 206.9 \\
\bottomrule
\end{tabular}
}
\vspace{-6mm}
\end{table*}
\subsection{Ablation Study}
Table~\ref{tab:ablation_models} analyzes the contributions of SFT initialization, RL data composition, and reward design. The ablation of reward model size and reward model selection are provided in Appendix~\ref{app:ablation_reward}.




\noindent\textbf{Effect of SFT Initialization.}
Replacing DRT-SFT with VisionR1-cold leads to a substantial increase in output token length and a modest accuracy drop. This suggests that structured initialization of DRT enables effective step-level supervision, whereas verbose and loosely organized long-CoT reasoning hinders reliable credit assignment during RL. Moreover, we find that VisionR1-dense (initialized with free-form CoT generated using the same teacher model, seed data, and response-length budgets as DRT-SFT) achieves lower accuracy with more output tokens than DRT, suggesting gains beyond generic concision or teacher-generated supervision alone (see Appendix~\ref{app:concise_cot} for details).

\noindent\textbf{Effect of RL Data Composition.}
Using vision-only data leads to weaker performance on challenging benchmarks such as LogicVista. This is consistent with our data design in Sec.~\ref{RL-dataset}.

\noindent\textbf{Effect of Reward Design.}
Using a standard GRPO objective leads to degraded performance, as the model tends to rely on shortcut strategies and answer guessing when answer-level signals fail to provide sufficient credit assignment for intermediate steps. Introducing step-level rewards alleviates this issue by enforcing alignment with reference reasoning trajectories and reducing hallucinated intermediate steps, but the model still exhibits limited exploration and tends to produce shallow reasoning chains. By rewarding non-trivial reasoning steps, the full reward design prevents premature convergence to trivial solutions and improves performance under constrained token budgets.

\begin{figure}[t]
    \centering
    \includegraphics[width=0.92\linewidth]{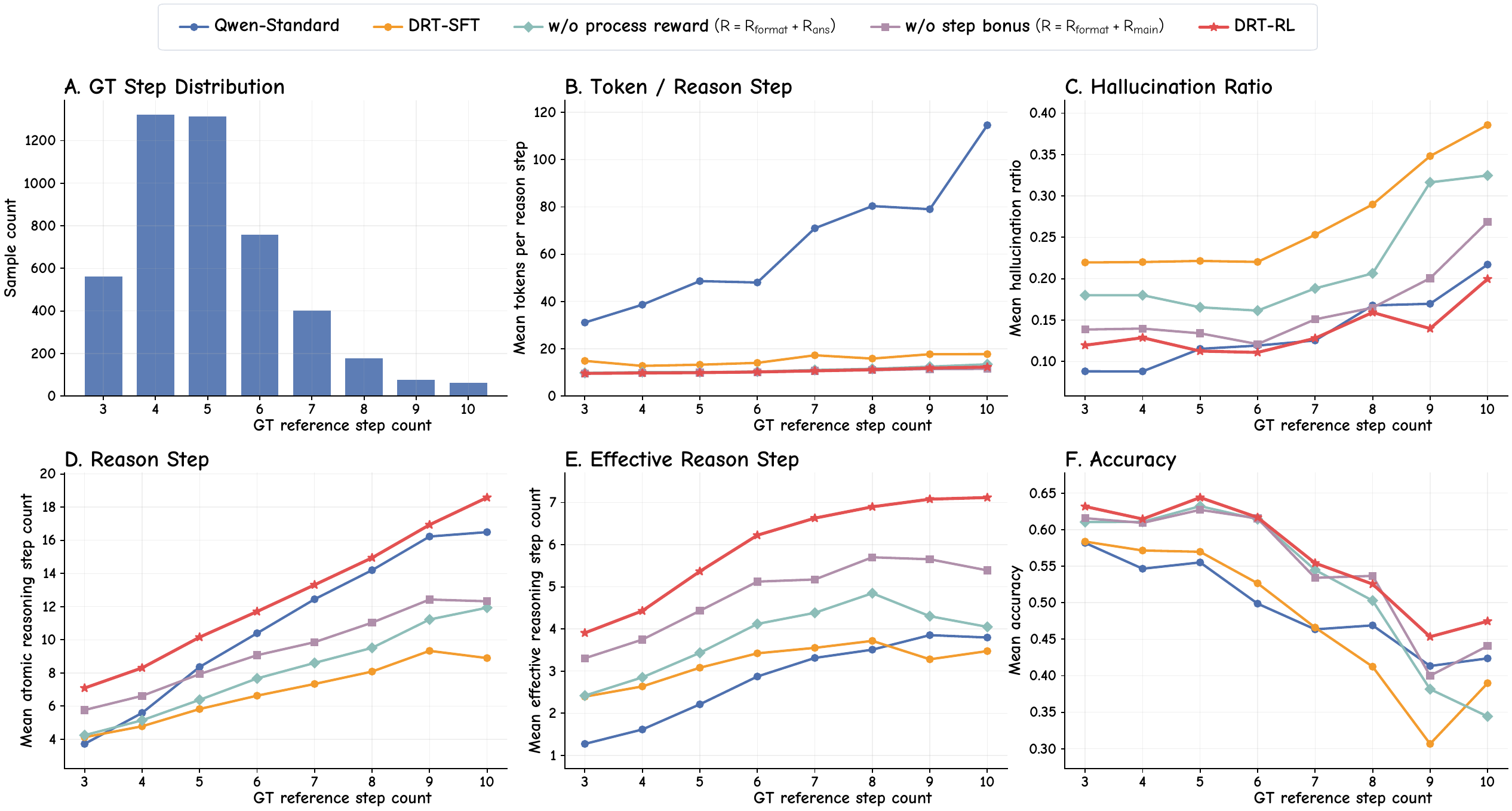}
    \vspace{-3mm}
    \caption{\textbf{Step-level analysis across reasoning difficulty using GPT-5.1 as an external judge.} Examples are grouped by ground-truth step counts, which serve as a proxy for reasoning difficulty. (A) shows the ground-truth step-count distribution, while (B)--(F) report per-bucket metrics: tokens per reasoning step, hallucination ratio, reasoning step, effective reasoning step, and accuracy.}
    \label{fig:mathvista_process_analysis}
    \vspace{-6mm}
\end{figure}

To better interpret ablation results and analyze how DRT improves reasoning efficiency and quality, we conduct an external step-level analysis of model predictions. Specifically, we construct reference step-level annotations for a combined subset of MathVista, MathVerse, and LogicVista using the pipeline described in Sec.~\ref{RL-dataset}, resulting in 4{,}671 evaluation samples. We group examples by the number of reference steps in the ground-truth trace, which serves as a proxy for problem difficulty, and report six complementary views: (A) \textit{GT Step Distribution}, the number of examples in each bucket; (B) \textit{Token / Reason Step}, the average number of generated tokens per \emph{predicted} reasoning step; (C) \textit{Hallucination Ratio}, the fraction of effective steps that rely on unsupported assumptions or fabricated visual evidence; (D) \textit{Reason Step}, the number of atomic reasoning steps; (E) \textit{Effective Reason Step}, the number of distinct solution-advancing steps; and (F) \textit{Accuracy}, the mean task accuracy in each bucket. Results are shown in Fig.~\ref{fig:mathvista_process_analysis} and evaluation prompts are provided in Appendix~\ref{app:step-level-gpt-judge}. 

\noindent\textbf{Step-level analysis.}
From Fig.~\ref{fig:mathvista_process_analysis}(B), Qwen-Standard (blue) exhibits a consistently high number of tokens per reasoning step, which increases sharply with problem difficulty, leading to inefficient and redundant reasoning. In contrast, DRT-SFT (orange) and its RL-enhanced variants maintain a significantly lower and more stable token-per-step ratio, benefiting from the structured reasoning paradigm of DRT. Notably, as shown in Fig.~\ref{fig:mathvista_process_analysis}(E), the total number of effective reasoning steps for DRT-SFT remains comparable to Qwen-Standard, indicating that the efficiency gains primarily arise from more compact reasoning within each step, rather than from reducing reasoning depth.

However, Fig.~\ref{fig:mathvista_process_analysis}(C) and (D) show that SFT alone, while enforcing a compact reasoning format, tends to reduce reasoning steps and increase the hallucination ratio. This results in insufficient effective reasoning steps in difficult tasks (Fig.~\ref{fig:mathvista_process_analysis}(E)) and degraded accuracy (Fig.~\ref{fig:mathvista_process_analysis}(F)). Similarly, RL with answer-only rewards (cyan) fails to address this issue due to the lack of step-level supervision. In contrast, incorporating step-level rewards (purple) significantly reduces hallucination, while the step bonus (red) further encourages sufficient reasoning depth on more complex instances. 
\section{Conclusion}
We introduce DRT, a paradigm for efficient and grounded multimodal reasoning. By disentangling visual grounding from logical deduction and compressing intermediate reasoning into compact traces, DRT mitigates the information dilution and weakened grounding often observed in long multimodal CoT reasoning.
To instantiate this paradigm, we first introduce Dense Trace Initialization to align the model with the DRT reasoning mode, improving token efficiency while preserving visual evidence. We further propose the Trace-Grounded Reinforcement Learning framework, which constructs grounded reference traces through a tri-perspective verification pipeline and performs Trace-Grounded GRPO to provide step-level supervision, encouraging faithful and logically grounded reasoning under tight token budgets.
Experiments across diverse benchmarks demonstrate that DRT achieves a significantly better efficiency-accuracy tradeoff than the Qwen3-VL baseline and prior efficient reasoning approaches, opening a more efficient path for next-generation MLLMs.


{\small
\bibliographystyle{plain}
\bibliography{Reference}

@article{bai2025qwen3,
  title={Qwen3-vl technical report},
  author={Bai, Shuai and Cai, Yuxuan and Chen, Ruizhe and Chen, Keqin and Chen, Xionghui and Cheng, Zesen and Deng, Lianghao and Ding, Wei and Gao, Chang and Ge, Chunjiang and others},
  journal={arXiv preprint arXiv:2511.21631},
  year={2025}
}

@inproceedings{wei2022chain,
 author = {Wei, Jason and Wang, Xuezhi and Schuurmans, Dale and Bosma, Maarten and ichter, brian and Xia, Fei and Chi, Ed and Le, Quoc V and Zhou, Denny},
 booktitle = {Advances in Neural Information Processing Systems},
 editor = {S. Koyejo and S. Mohamed and A. Agarwal and D. Belgrave and K. Cho and A. Oh},
 pages = {24824--24837},
 publisher = {Curran Associates, Inc.},
 title = {Chain-of-Thought Prompting Elicits Reasoning in Large Language Models},
 url = {https://proceedings.neurips.cc/paper_files/paper/2022/file/9d5609613524ecf4f15af0f7b31abca4-Paper-Conference.pdf},
 volume = {35},
 year = {2022}
}

@article{guo2025deepseek,
  title={Deepseek-r1: Incentivizing reasoning capability in llms via reinforcement learning},
  author={Guo, Daya and Yang, Dejian and Zhang, Haowei and Song, Junxiao and Wang, Peiyi and Zhu, Qihao and Xu, Runxin and Zhang, Ruoyu and Ma, Shirong and Bi, Xiao and others},
  journal={arXiv preprint arXiv:2501.12948},
  year={2025}
}

@misc{anthropic_claude_opus_4_6,
  author       = {Anthropic},
  title        = {Claude Opus 4.6},
  year         = {2026},
  url          = {https://www.anthropic.com/news/claude-opus-4-6},
  howpublished = {\url{https://www.anthropic.com/news/claude-opus-4-6}}
}

@misc{OpenAI_o3,
  author       = {OpenAI},
  title        = {OpenAI o3},
  year         = {2025},
  url          = {https://openai.com/index/introducing-o3-and-o4-mini/},
  howpublished = {\url{https://openai.com/index/introducing-o3-and-o4-mini/}}
}

@inproceedings{huang2025vision,
title={Vision-R1: Incentivizing Reasoning Capability in Multimodal Large Language Models},
author={Wenxuan Huang and Bohan Jia and Shaosheng Cao and Zheyu Ye and Fei zhao and Zhe Xu and Yao Hu and Shaohui Lin},
booktitle={The Fourteenth International Conference on Learning Representations},
year={2026},
url={https://openreview.net/forum?id=UZIjskfbfU}
}

@inproceedings{lu2024mathvista,
title={MathVista: Evaluating Mathematical Reasoning of Foundation Models in Visual Contexts},
author={Pan Lu and Hritik Bansal and Tony Xia and Jiacheng Liu and Chunyuan Li and Hannaneh Hajishirzi and Hao Cheng and Kai-Wei Chang and Michel Galley and Jianfeng Gao},
booktitle={The Twelfth International Conference on Learning Representations},
year={2024},
url={https://openreview.net/forum?id=KUNzEQMWU7}
}

@inproceedings{zhang2024mathverse,
  title={Mathverse: Does your multi-modal llm truly see the diagrams in visual math problems?},
  author={Zhang, Renrui and Jiang, Dongzhi and Zhang, Yichi and Lin, Haokun and Guo, Ziyu and Qiu, Pengshuo and Zhou, Aojun and Lu, Pan and Chang, Kai-Wei and Qiao, Yu and others},
  booktitle={European Conference on Computer Vision},
  pages={169--186},
  year={2024},
  organization={Springer}
}

@article{xiao2024logicvista,
  title={Logicvista: Multimodal llm logical reasoning benchmark in visual contexts},
  author={Xiao, Yijia and Sun, Edward and Liu, Tianyu and Wang, Wei},
  journal={arXiv preprint arXiv:2407.04973},
  year={2024}
}

@article{cheng2025video_holmes,
  title={Video-Holmes: Can MLLM Think Like Holmes for Complex Video Reasoning?},
  author={Cheng, Junhao and Ge, Yuying and Wang, Teng and Ge, Yixiao and Liao, Jing and Shan, Ying},
  journal={arXiv preprint arXiv:2505.21374},
  year={2025}
}

@inproceedings{snell2025scaling,
title={Scaling {LLM} Test-Time Compute Optimally Can be More Effective than Scaling Parameters for Reasoning},
author={Charlie Victor Snell and Jaehoon Lee and Kelvin Xu and Aviral Kumar},
booktitle={The Thirteenth International Conference on Learning Representations},
year={2025},
url={https://openreview.net/forum?id=4FWAwZtd2n}
}

@article{li2025compressing,
  title={Compressing chain-of-thought in llms via step entropy},
  author={Li, Zeju and Zhong, Jianyuan and Zheng, Ziyang and Wen, Xiangyu and Xu, Zhijian and Cheng, Yingying and Zhang, Fan and Xu, Qiang},
  journal={arXiv preprint arXiv:2508.03346},
  year={2025}
}

@inproceedings{she2025hawkeye,
  title={Hawkeye: Model Collaboration for Efficient Reasoning},
  author={She, Jianshu and Li, Zhuohao and Huang, Zhemin and Li, Qi and Xu, Peiran and Li, Haonan and Ho, Qirong},
  booktitle={Second Conference on Language Modeling},
  year={2025}
}

@inproceedings{huang2024opera,
  title={Opera: Alleviating hallucination in multi-modal large language models via over-trust penalty and retrospection-allocation},
  author={Huang, Qidong and Dong, Xiaoyi and Zhang, Pan and Wang, Bin and He, Conghui and Wang, Jiaqi and Lin, Dahua and Zhang, Weiming and Yu, Nenghai},
  booktitle={Proceedings of the IEEE/CVF Conference on Computer Vision and Pattern Recognition},
  pages={13418--13427},
  year={2024}
}

@inproceedings{li2024chain,
title={Chain of Thought Empowers Transformers to Solve Inherently Serial Problems},
author={Zhiyuan Li and Hong Liu and Denny Zhou and Tengyu Ma},
booktitle={The Twelfth International Conference on Learning Representations},
year={2024},
url={https://openreview.net/forum?id=3EWTEy9MTM}
}

@inproceedings{yao2023tree,
 author = {Yao, Shunyu and Yu, Dian and Zhao, Jeffrey and Shafran, Izhak and Griffiths, Tom and Cao, Yuan and Narasimhan, Karthik},
 booktitle = {Advances in Neural Information Processing Systems},
 editor = {A. Oh and T. Naumann and A. Globerson and K. Saenko and M. Hardt and S. Levine},
 pages = {11809--11822},
 publisher = {Curran Associates, Inc.},
 title = {Tree of Thoughts: Deliberate Problem Solving with Large Language Models},
 url = {https://proceedings.neurips.cc/paper_files/paper/2023/file/271db9922b8d1f4dd7aaef84ed5ac703-Paper-Conference.pdf},
 volume = {36},
 year = {2023}
}

@inproceedings{besta2024graph,
  title={Graph of thoughts: Solving elaborate problems with large language models},
  author={Besta, Maciej and Blach, Nils and Kubicek, Ales and Gerstenberger, Robert and Podstawski, Michal and Gianinazzi, Lukas and Gajda, Joanna and Lehmann, Tomasz and Niewiadomski, Hubert and Nyczyk, Piotr and others},
  booktitle={Proceedings of the AAAI conference on artificial intelligence},
  volume={38},
  number={16},
  pages={17682--17690},
  year={2024}
}

@inproceedings{chen2024alphamath,
 author = {Chen, Guoxin and Liao, Minpeng and Li, Chengxi and Fan, Kai},
 booktitle = {Advances in Neural Information Processing Systems},
 doi = {10.52202/079017-0870},
 editor = {A. Globerson and L. Mackey and D. Belgrave and A. Fan and U. Paquet and J. Tomczak and C. Zhang},
 pages = {27689--27724},
 publisher = {Curran Associates, Inc.},
 title = {AlphaMath Almost Zero: Process Supervision without Process},
 url = {https://proceedings.neurips.cc/paper_files/paper/2024/file/30dfe47a3ccbee68cffa0c19ccb1bc00-Paper-Conference.pdf},
 volume = {37},
 year = {2024}
}

@inproceedings{feng2023towards,
 author = {Feng, Guhao and Zhang, Bohang and Gu, Yuntian and Ye, Haotian and He, Di and Wang, Liwei},
 booktitle = {Advances in Neural Information Processing Systems},
 editor = {A. Oh and T. Naumann and A. Globerson and K. Saenko and M. Hardt and S. Levine},
 pages = {70757--70798},
 publisher = {Curran Associates, Inc.},
 title = {Towards Revealing the Mystery behind Chain of Thought: A Theoretical Perspective},
 url = {https://proceedings.neurips.cc/paper_files/paper/2023/file/dfc310e81992d2e4cedc09ac47eff13e-Paper-Conference.pdf},
 volume = {36},
 year = {2023}
}

@article{xu2025cod,
    title={Chain of Draft: Thinking Faster by Writing Less},
    author={Xu, Silei and Xie, Wenhao and Zhao, Lingxiao and He, Pengcheng},
    journal={arXiv preprint arXiv:2502.18600},
    year={2025}
}

@inproceedings{xia2025tokenskip,
  title={Tokenskip: Controllable chain-of-thought compression in llms},
  author={Xia, Heming and Leong, Chak Tou and Wang, Wenjie and Li, Yongqi and Li, Wenjie},
  booktitle={Proceedings of the 2025 Conference on Empirical Methods in Natural Language Processing},
  pages={3351--3363},
  year={2025}
}

@article{jiang2025drp,
  title={DRP: Distilled Reasoning Pruning with Skill-aware Step Decomposition for Efficient Large Reasoning Models},
  author={Jiang, Yuxuan and Li, Dawei and Ferraro, Frank},
  journal={arXiv preprint arXiv:2505.13975},
  year={2025}
}

@article{li2025thinkless,
  title={Thinkless: A training-free inference-efficient method for reducing reasoning redundancy},
  author={Li, Gengyang and Gao, Yifeng and Li, Yuming and Wu, Yunfang},
  journal={arXiv preprint arXiv:2505.15684},
  year={2025}
}

@article{li2025selfbudgeter,
  title={Selfbudgeter: Adaptive token allocation for efficient llm reasoning},
  author={Li, Zheng and Dong, Qingxiu and Ma, Jingyuan and Zhang, Di and Jia, Kai and Sui, Zhifang},
  journal={arXiv preprint arXiv:2505.11274},
  year={2025}
}

@article{nohara2026optimal,
  title={On the Optimal Reasoning Length for RL-Trained Language Models},
  author={Nohara, Daisuke and Nakamura, Taishi and Yokota, Rio},
  journal={arXiv preprint arXiv:2602.09591},
  year={2026}
}

@inproceedings{yang2025visionthink,
title={VisionThink: Smart and Efficient Vision Language Model via Reinforcement Learning},
author={Senqiao Yang and Junyi Li and Xin Lai and Jinming Wu and Wei Li and Zejun MA and Bei Yu and Hengshuang Zhao and Jiaya Jia},
booktitle={The Thirty-ninth Annual Conference on Neural Information Processing Systems},
year={2025},
url={https://openreview.net/forum?id=R6m6bNnmWm}
}

@article{zhang2026chain,
  title={Chain-of-thought compression should not be blind: V-skip for efficient multimodal reasoning via dual-path anchoring},
  author={Zhang, Dongxu and Sun, Yiding and Tan, Cheng and Yan, Wenbiao and Yang, Ning and Zhu, Jihua and Zhang, Haijun},
  journal={arXiv preprint arXiv:2601.13879},
  year={2026}
}

@inproceedings{NEURIPS2023_e6c2e85d,
 author = {Li, Huayang and Lan, Tian and Fu, Zihao and Cai, Deng and Liu, Lemao and Collier, Nigel and Watanabe, Taro and Su, Yixuan},
 booktitle = {Advances in Neural Information Processing Systems},
 editor = {A. Oh and T. Naumann and A. Globerson and K. Saenko and M. Hardt and S. Levine},
 pages = {72888--72903},
 publisher = {Curran Associates, Inc.},
 title = {Repetition In Repetition Out: Towards Understanding Neural Text Degeneration from the Data Perspective},
 url = {https://proceedings.neurips.cc/paper_files/paper/2023/file/e6c2e85db1f1039177c4495ccd399ac4-Paper-Conference.pdf},
 volume = {36},
 year = {2023}
}

@inproceedings{yao2025understanding,
  title={Understanding the repeat curse in large language models from a feature perspective},
  author={Yao, Junchi and Yang, Shu and Xu, Jianhua and Hu, Lijie and Li, Mengdi and Wang, Di},
  booktitle={Findings of the Association for Computational Linguistics: ACL 2025},
  pages={7787--7815},
  year={2025}
}

@misc{singh2025openai,
  author       = {OpenAI},
  title        = {{GPT-5.1 Instant and GPT-5.1 Thinking System Card Addendum}},
  year         = {2025},
  howpublished = {\url{https://deploymentsafety.openai.com/gpt-5-1}}
}

@article{ouyang2022training,
  title={Training language models to follow instructions with human feedback},
  author={Ouyang, Long and Wu, Jeffrey and Jiang, Xu and Almeida, Diogo and Wainwright, Carroll and Mishkin, Pamela and Zhang, Chong and Agarwal, Sandhini and Slama, Katarina and Ray, Alex and others},
  journal={Advances in neural information processing systems},
  volume={35},
  pages={27730--27744},
  year={2022}
}

@inproceedings{wang2023self,
  title={Self-instruct: Aligning language models with self-generated instructions},
  author={Wang, Yizhong and Kordi, Yeganeh and Mishra, Swaroop and Liu, Alisa and Smith, Noah A and Khashabi, Daniel and Hajishirzi, Hannaneh},
  booktitle={Proceedings of the 61st annual meeting of the association for computational linguistics (volume 1: long papers)},
  pages={13484--13508},
  year={2023}
}

@article{cobbe2021training,
  title={Training verifiers to solve math word problems},
  author={Cobbe, Karl and Kosaraju, Vineet and Bavarian, Mohammad and Chen, Mark and Jun, Heewoo and Kaiser, Lukasz and Plappert, Matthias and Tworek, Jerry and Hilton, Jacob and Nakano, Reiichiro and others},
  journal={arXiv preprint arXiv:2110.14168},
  year={2021}
}

@inproceedings{
yu2025dapo,
title={{DAPO}: An Open-Source {LLM} Reinforcement Learning System at Scale},
author={Qiying Yu and Zheng Zhang and Ruofei Zhu and Yufeng Yuan and Xiaochen Zuo and YuYue and Weinan Dai and Tiantian Fan and Gaohong Liu and Juncai Liu and LingJun Liu and Xin Liu and Haibin Lin and Zhiqi Lin and Bole Ma and Guangming Sheng and Yuxuan Tong and Chi Zhang and Mofan Zhang and Ru Zhang and Wang Zhang and Hang Zhu and Jinhua Zhu and Jiaze Chen and Jiangjie Chen and Chengyi Wang and Hongli Yu and Yuxuan Song and Xiangpeng Wei and Hao Zhou and Jingjing Liu and Wei-Ying Ma and Ya-Qin Zhang and Lin Yan and Yonghui Wu and Mingxuan Wang},
booktitle={The Thirty-ninth Annual Conference on Neural Information Processing Systems},
year={2025},
url={https://openreview.net/forum?id=2a36EMSSTp}
}

@inproceedings{kwon2023efficient,
  title={Efficient Memory Management for Large Language Model Serving with PagedAttention},
  author={Woosuk Kwon and Zhuohan Li and Siyuan Zhuang and Ying Sheng and Lianmin Zheng and Cody Hao Yu and Joseph E. Gonzalez and Hao Zhang and Ion Stoica},
  booktitle={Proceedings of the ACM SIGOPS 29th Symposium on Operating Systems Principles},
  year={2023}
}

@inproceedings{kingma2015adam,
  author    = {Kingma, Diederik P. and Ba, Jimmy},
  title     = {Adam: A Method for Stochastic Optimization},
  booktitle = {International Conference on Learning Representations (ICLR)},
  year      = {2015},
  url       = {https://arxiv.org/abs/1412.6980}
}

@misc{ms-swift,
      title={SWIFT:A Scalable lightWeight Infrastructure for Fine-Tuning},
      author={Yuze Zhao and Jintao Huang and Jinghan Hu and Xingjun Wang and Yunlin Mao and Daoze Zhang and Zeyinzi Jiang and Zhikai Wu and Baole Ai and Ang Wang and Wenmeng Zhou and Yingda Chen},
      year={2024},
      eprint={2408.05517},
      archivePrefix={arXiv},
      primaryClass={cs.CL},
      url={https://arxiv.org/abs/2408.05517},
}

@article{sheng2024hybridflow,
  title   = {HybridFlow: A Flexible and Efficient RLHF Framework},
  author  = {Guangming Sheng and Chi Zhang and Zilingfeng Ye and Xibin Wu and Wang Zhang and Ru Zhang and Yanghua Peng and Haibin Lin and Chuan Wu},
  year    = {2024},
  journal = {arXiv preprint arXiv: 2409.19256}
}

@inproceedings{lightman2023let,
  title={Let's verify step by step},
  author={Lightman, Hunter and Kosaraju, Vineet and Burda, Yuri and Edwards, Harrison and Baker, Bowen and Lee, Teddy and Leike, Jan and Schulman, John and Sutskever, Ilya and Cobbe, Karl},
  booktitle={The twelfth international conference on learning representations},
  year={2023}
}

@inproceedings{
setlur2025rewarding,
title={Rewarding Progress: Scaling Automated Process Verifiers for {LLM} Reasoning},
author={Amrith Setlur and Chirag Nagpal and Adam Fisch and Xinyang Geng and Jacob Eisenstein and Rishabh Agarwal and Alekh Agarwal and Jonathan Berant and Aviral Kumar},
booktitle={The Thirteenth International Conference on Learning Representations},
year={2025},
url={https://openreview.net/forum?id=A6Y7AqlzLW}
}

@article{yang2025qwen3,
  title={Qwen3 technical report},
  author={Yang, An and Li, Anfeng and Yang, Baosong and Zhang, Beichen and Hui, Binyuan and Zheng, Bo and Yu, Bowen and Gao, Chang and Huang, Chengen and Lv, Chenxu and others},
  journal={arXiv preprint arXiv:2505.09388},
  year={2025}
}

@article{huang2026learning,
  title={Learning Adaptive Reasoning Paths for Efficient Visual Reasoning},
  author={Huang, Yixu and Zhu, Tinghui and Chen, Muhao},
  journal={arXiv preprint arXiv:2604.14568},
  year={2026}
}

@inproceedings{hao2025training,
title={Training Large Language Models to Reason in a Continuous Latent Space},
author={Shibo Hao and Sainbayar Sukhbaatar and DiJia Su and Xian Li and Zhiting Hu and Jason E Weston and Yuandong Tian},
booktitle={Second Conference on Language Modeling},
year={2025},
url={https://openreview.net/forum?id=Itxz7S4Ip3}
}

@inproceedings{shen2025codi,
  title={Codi: Compressing chain-of-thought into continuous space via self-distillation},
  author={Shen, Zhenyi and Yan, Hanqi and Zhang, Linhai and Hu, Zhanghao and Du, Yali and He, Yulan},
  booktitle={Proceedings of the 2025 Conference on Empirical Methods in Natural Language Processing},
  pages={677--693},
  year={2025}
}

@inproceedings{xu2025softcot,
  title={Softcot: Soft chain-of-thought for efficient reasoning with llms},
  author={Xu, Yige and Guo, Xu and Zeng, Zhiwei and Miao, Chunyan},
  booktitle={Proceedings of the 63rd Annual Meeting of the Association for Computational Linguistics (Volume 1: Long Papers)},
  pages={23336--23351},
  year={2025}
}

@article{wei2025sim,
  title={SIM-CoT: Supervised Implicit Chain-of-Thought},
  author={Wei, Xilin and Liu, Xiaoran and Zang, Yuhang and Dong, Xiaoyi and Cao, Yuhang and Wang, Jiaqi and Qiu, Xipeng and Lin, Dahua},
  journal={arXiv preprint arXiv:2509.20317},
  year={2025}
}

@article{jeon2026vision,
  title={Vision-aligned Latent Reasoning for Multi-modal Large Language Model},
  author={Jeon, Byungwoo and Jeong, Yoonwoo and Lee, Hyunseok and Cho, Minsu and Shin, Jinwoo},
  journal={arXiv preprint arXiv:2602.04476},
  year={2026}
}
}

\clearpage
\appendix

\section{Quantitative Accounting of The End-to-end Costs}\label{app:all_cost}
We performed an end-to-end accounting of the API and accelerator costs of the reported pipeline. Dataset generation and verification consumed approximately 200M GPT-5.1 input tokens and 42M output tokens, corresponding to approximately 670 USD at the current standard API price. SFT consumed 1,600 H100 GPU-hours, while RL consumed 1,920 A100 GPU-hours. The reported 1,920 A100 GPU-hours include policy rollout, local reward-judge inference, and policy optimization; therefore, no additional reward-model API or accelerator cost is omitted. Using current publicly listed AWS prices, the total one-time cost is approximately 13,200 USD (8xA100 17.712 USD/hour, 8xH100 41.528 USD/hour, based on current AWS EC2 Capacity Blocks pricing).

Under our measured single-A100 serving results, processing one million requests requires 186.4 A100 GPU-hours with Qwen-Standard but only 40.2 A100 GPU-hours with DRT-RL, saving 146.2 A100 GPU-hours per million requests. Consequently, the complete upfront cost is amortized after approximately 41 million requests. As an illustrative high-volume deployment, a service requiring 200 A100 GPUs to sustain the Qwen-Standard throughput of 298 QPS would require only approximately 44 A100 GPUs with DRT-RL. Under the same public cloud prices, the upfront cost would therefore be recovered in approximately 1.6 days of sustained operation.

\section{Reward and Judge Configuration}\label{app:verl_hyperparameters}
The reward judge is \texttt{Qwen3-235B-A22B-Instruct-2507}~\cite{yang2025qwen3} served by vLLM~\cite{kwon2023efficient}.
We use deterministic decoding (\texttt{temperature}=0) for the reward judge.

\noindent\textbf{Format reward.}
We implement the format reward using rule-based checks to enforce the DRT reasoning structure. Specifically, trajectories that contain well-formed \texttt{<think>} and \texttt{<answer>} segments receive a small positive reward of $+0.1$, while invalid formats are assigned a penalty of $-1.0$. To encourage concise and high-density reasoning, we further segment the \texttt{<think>} content using symbolic connectors (e.g., $\rightarrow$). If any segment exceeds a maximum length of 15 tokens, a penalty of $-0.05$ is applied per violation, capped at $-0.25$ in total. These coefficients are intentionally conservative: the format reward primarily serves as a structural regularizer to preserve the DRT reasoning pattern learned during SFT, with penalties mainly applied to prevent format violations and overly verbose reasoning.

\noindent\textbf{Main reward.}
The main reward prioritizes final answer correctness while providing partial credit for intermediate reasoning. Specifically, the answer weight $\lambda_{\text{main}}$ is set to 0.9, and the partial-credit weight $\lambda_{\text{proc}}$ is 0.8, ensuring that correct answers dominate the reward signal while still allowing meaningful gradients from partially correct reasoning trajectories. 
To penalize ungrounded reasoning, we apply a multiplicative hallucination correction with penalty ratio $\lambda_{\text{hall}} = 0.5$, which moderately down-weights answers supported by hallucinated intermediate steps without overly discouraging early-stage exploration.

\noindent\textbf{Step bonus reward.}
The step bonus reward encourages effective multi-step reasoning and controlled exploration. The step bonus weight $\lambda_{\text{step}}$ is set to 0.3, and the deep-exploration bonus weight $\lambda_{\text{explore}}$ is 0.12, with a cap $N_{\text{cap}} = 4$ to prevent excessive reward from overly long reasoning chains. 
For difficulty-aware exploration, we use threshold/cap values of 6/9 for Vision-R1-RL and 10/18 for DAPO. The difficulty factor $\rho_{\text{diff}}$ is set to 0 when the number of reference steps $|S^*|$ is below the threshold, increases linearly from 0 to 1 as $|S^*|$ grows to the cap, and saturates thereafter. This design activates exploration incentives primarily on more complex problems, aligning reward shaping with reasoning difficulty.

\section{Detailed Training Configuration}\label{app:training}
\paragraph{SFT Details.}
We perform full-parameter fine-tuning without freezing the language model.
The visual encoder is kept frozen, while the multimodal aligner remains trainable, with a reduced learning rate multiplier of 0.2 applied to the vision branch.
We use Adam~\cite{kingma2015adam} with zero weight decay and a cosine learning rate schedule, with a peak learning rate of $6.0 \times 10^{-6}$ and a minimum of $1.0 \times 10^{-6}$.
Training is conducted with a micro-batch size of 1 and a global batch size of 32, using dynamic batching with a maximum sequence length of 16{,}384 tokens.
We employ Flash Attention for efficiency, together with tensor parallelism of 8 and context parallelism of 4.
Our implementation is based on the \texttt{ms-swift} framework~\cite{ms-swift} built on Megatron-LM with DeepSpeed support, with activation recomputation and other memory-efficient optimizations enabled.
We train for up to 2{,}000 iterations, with early stopping applied at 800 iterations, resulting in 369{,}455 training samples processed. Training is performed on 32 H100-SXM-80GB GPUs, taking approximately 50 hours to complete.
\paragraph{RL Details.}
Both the trainable policy and the frozen reference model are initialized from the SFT checkpoint at 800 iterations.
We adopt GRPO~\cite{guo2025deepseek} with group-based sampling ($G=16$) and KL regularization ($\beta = 0.01$) with respect to the frozen reference model, without entropy regularization.
Optimization uses Adam~\cite{kingma2015adam} with a learning rate of $1.0 \times 10^{-6}$, a training batch size of 512, a PPO mini-batch size of 128, and a micro-batch size of 1 per GPU.
The maximum prompt and response lengths are both set to 2{,}048 tokens.
The reward model supports a maximum context length of 12{,}800 tokens, with reward-side prompt and response limits of 12{,}000 and 512 tokens, respectively.
Training is implemented using Megatron-style parallelism, where both the trainable policy and the frozen reference model use tensor, pipeline, and context parallelism sizes of 2, 2, and 1, respectively.
Policy rollout generation is accelerated using vLLM~\cite{kwon2023efficient} with tensor parallelism size 4, and RL training is implemented with verl~\cite{sheng2024hybridflow}. Training is performed on 16 A100-SXM-80GB GPUs, taking approximately 120 hours to complete.

\begin{figure*}[t]
    \centering
    \includegraphics[width=\textwidth]{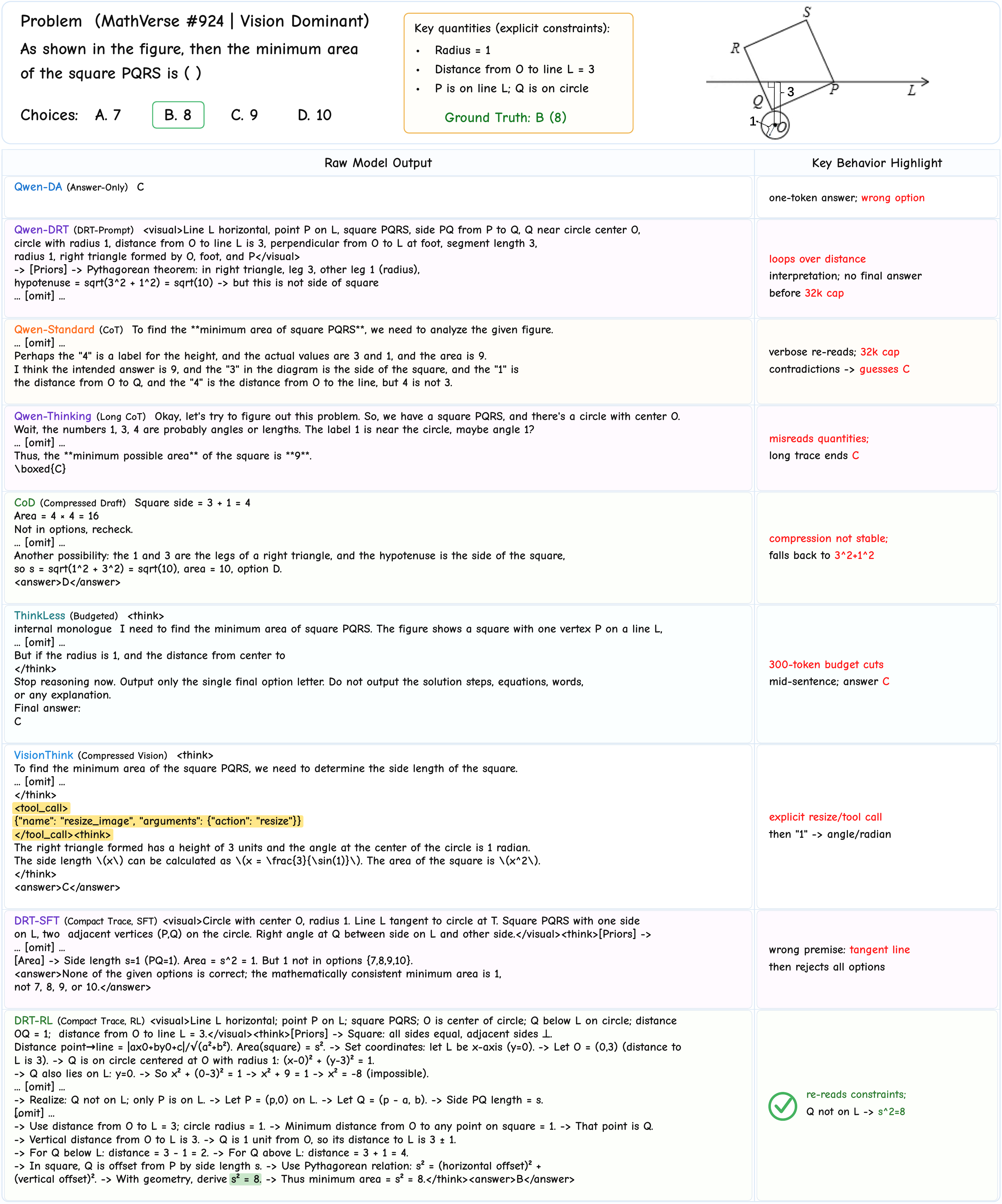}
    \caption{\textbf{Case-level comparison of reasoning behaviors on a MathVerse geometry problem.}}
    \label{fig:main_result_showcase}
    \vspace{-4mm}
\end{figure*}

\section{Zero-Shot DRT Prompting on Stronger Models}\label{app:zero_shot}
We conduct zero-shot prompting experiments on stronger general-purpose models to assess whether DRT prompting improves their reasoning performance without DRT-specific training. We evaluate GPT-5.5 and Gemini-2.5-Flash, for which our available APIs permit comparable evaluation of visible reasoning outputs; Gemini-2.5-Pro/Gemini-3 could not be configured to disable built-in thinking in our setup, and we lacked reliable Claude API access. Standard CoT follows the default prompting format while concise CoT and DRT prompt the models to generate concise CoT-style or DRT-style reasoning traces.

\begin{table}[ht]
    \centering
    \caption{
        Zero-shot prompting results on stronger models.
    }
    \label{tab:stronger_models_prompting}
    \small
    \setlength{\tabcolsep}{3pt}
    \renewcommand{\arraystretch}{1.15}
    \resizebox{\linewidth}{!}{%
        \begin{tabular}{lcccccccc}
            \toprule
            \textbf{Model}
            & \textbf{Prompt}
            & \textbf{MathVista}
            & \textbf{MathVerse}
            & \textbf{LogicVista}
            & \textbf{Video-Holmes}
            & \textbf{GSM8K}
            & \textbf{Avg. Acc.$\uparrow$}
            & \textbf{Avg. Tokens$\downarrow$} \\
            \midrule
            GPT-5.5
            & Standard CoT
            & 72.60
            & 60.25
            & 66.44
            & {54.11}
            & 94.24
            & 69.53
            & {66.7} \\

            GPT-5.5
            & Concise CoT
            & 79.40
            & 80.51
            & 74.27
            & 53.57
            & {96.51}
            & 76.85
            & 97.4 \\

            GPT-5.5
            & DRT
            & {81.70}
            & {81.83}
            & {76.96}
            & 52.69
            & 95.68
            & {77.77}
            & 130.1 \\

            Gemini-2.5-Flash
            & Standard CoT
            & 74.80
            & 75.18
            & 66.67
            & 50.52
            & 88.93
            & 71.22
            & 1087.3 \\

            Gemini-2.5-Flash
            & Concise CoT
            & {76.90}
            & {76.37}
            & {71.14}
            & 49.16
            & 88.02
            & 72.32
            & 1100.2 \\

            Gemini-2.5-Flash
            & DRT
            & 75.50
            & 75.53
            & 69.80
            & {50.57}
            & {94.62}
            & {73.20}
            & {915.4} \\
            \bottomrule
        \end{tabular}%
    }
\end{table}

DRT achieves the highest average accuracy for both models. Using concise CoT as the stronger prompting baseline, DRT improves accuracy by +0.92 and +0.88 points, respectively. This provides evidence that the DRT representation itself contributes beyond the proposed SFT/RL pipeline. However, prompting alone does not consistently realize DRT's intended token efficiency. GPT-5.5 uses 33.5\% more tokens than concise CoT, suggesting that stronger models already exhibit relatively efficient reasoning behavior under standard prompting, while the unfamiliar DRT format mainly encourages more explicit reasoning rather than reducing redundancy. In contrast, Gemini-2.5-Flash shows behavior closer to our Qwen3-VL-8B results: DRT reduces mean tokens by 16.8\%, but occasional repetitive verification and backtracking still lead to a heavy-tailed length distribution. These observations suggest that reasoning redundancy can be more prominent in models without highly optimized reasoning behaviors, prompting alone is insufficient to reliably shift the underlying reasoning behavior or achieve substantial output token reduction, and DRT can provide larger efficiency gains after training. Overall, zero-shot DRT validates the effectiveness of the proposed reasoning structure, while DRT-specific SFT/RL is important for internalizing a stable, high-density, and token-efficient reasoning mode.

\section{Case-level Analysis of Reasoning Behaviors}\label{app:case_study}
To better understand the aggregate trends in Table~\ref{tab:main_results}, we present a qualitative case study in Figure~\ref{fig:main_result_showcase}. The example is a MathVerse geometry problem that requires recognizing the explicit constraints from the diagram, including the circle radius, the distance from the circle center to the line, and the fact that only point $P$ lies on line $L$ while $Q$ lies on the circle.

The case illustrates several representative failure modes of existing reasoning strategies. Qwen-DA produces a very short answer but selects the wrong option. Prompt-only Qwen-DRT starts from the intended structured format but drifts into repeated self-correction and fails to produce a valid final answer before the generation limit. Qwen-Standard produces a long chain-of-thought trace with repeated diagram re-interpretations and eventually guesses the wrong option. CoD initially compresses reasoning but later expands into incompatible interpretations, showing a long-tail failure mode. Qwen-Thinking shows a similar issue to Qwen-Standard, producing an overly long and low-density reasoning trace that repeatedly re-parses the diagram but still reaches the wrong answer. Under a 300-token budget, ThinkLess fails to complete its reasoning process, resulting in truncated outputs. VisionThink triggers an additional image-resizing call, suggesting difficulty in reliably extracting information from the visual input. DRT-SFT produces a compact structured trace but relies on an incorrect tangent-line premise.

In contrast, DRT-RL keeps the reasoning trace relatively concise while preserving the key correction step: it first rejects inconsistent placements where $Q$ is assumed to lie on line $L$, then uses the corrected constraint that only $P$ lies on $L$ and derives the correct answer. This case supports the main finding that DRT-RL improves reasoning quality while maintaining a compact and controllable output format.

\begin{table}[t]
\centering
\caption{System-level efficiency comparison across efficient reasoning methods on five reasoning benchmarks. For each benchmark, we report throughput (QPS$\uparrow$) and latency (Latency$\downarrow$). The final columns report the macro average of QPS and Latency across all five benchmarks.}
\resizebox{1.0\linewidth}{!}{
\begin{tabular}{lcccccccccccc}
\toprule
\multirow{2}{*}{\textbf{Method}}
& \multicolumn{2}{c}{\textbf{MathVista}}
& \multicolumn{2}{c}{\textbf{MathVerse}}
& \multicolumn{2}{c}{\textbf{LogicVista}}
& \multicolumn{2}{c}{\textbf{GSM8K}}
& \multicolumn{2}{c}{\textbf{Video-Holmes}}
& \multicolumn{2}{c}{\textbf{AVG.}} \\
\cmidrule(lr){2-3} \cmidrule(lr){4-5} \cmidrule(lr){6-7} \cmidrule(lr){8-9} \cmidrule(lr){10-11} \cmidrule(lr){12-13}
& QPS$\uparrow$ & Latency$\downarrow$
& QPS$\uparrow$ & Latency$\downarrow$
& QPS$\uparrow$ & Latency$\downarrow$
& QPS$\uparrow$ & Latency$\downarrow$
& QPS$\uparrow$ & Latency$\downarrow$
& QPS$\uparrow$ & Latency$\downarrow$ \\
\midrule

Qwen-DA
& 13.56 & 9.07
& 14.94 & 8.39
& 22.82 & 4.35
& 46.19 & 2.42
& 2.69 & 44.93
& 20.04 & 13.83 \\

Qwen-DRT
& 1.42 & 23.37
& 0.97 & 60.12
& 0.74 & 37.20
& 4.29 & 3.33
& 1.76 & 24.87
& 1.84 & 29.78 \\

Qwen-Standard
& 1.22 & 32.75
& 0.99 & 59.08
& 0.31 & 160.29
& 3.33 & 7.83
& 1.62 & 26.74
& 1.49 & 57.34 \\

Qwen-Thinking
& 0.68 & 75.22
& 0.51 & 119.89
& 0.29 & 187.64
& 1.47 & 32.65
& 0.51 & 117.45
& 0.69 & 106.57 \\

CoD
& 2.29 & 10.30
& 1.67 & 32.38
& 1.58 & 6.23
& 5.13 & 2.09
& 1.69 & 26.77
& 2.47 & 15.55 \\

ThinkLess
& 5.36 & 11.72
& 5.88 & 10.82
& 6.10 & 10.08
& 10.42 & 5.99
& 1.47 & 42.56
& 5.84 & 16.23 \\

VisionThink
& 1.04 & 48.01
& 0.71 & 56.56
& 0.85 & 61.25
& 2.21 & 19.58
& 0.39 & 140.01
& 1.04 & 65.08 \\

DRT-SFT
& 6.14 & 9.62
& 6.43 & 9.87
& 7.91 & 7.45
& 16.94 & 3.67
& 2.75 & 22.18
& 8.03 & 10.56 \\

DRT-RL
& 5.43 & 10.97
& 5.40 & 11.67
& 6.83 & 8.66
& 14.17 & 4.39
& 2.73 & 22.40
& 6.91 & 11.62 \\

\bottomrule
\end{tabular}
}
\label{tab:main_latency}
\vspace{-5mm}
\end{table}

\section{System-Level Efficiency Evaluation}
\subsection{Evaluation Setting}\label{app:efficient_eval_setting}
All system-level efficiency results are measured using the vLLM framework~\cite{kwon2023efficient} on a single NVIDIA A100-SXM4-80GB GPU under an autoregressive decoding setup. We evaluate all methods with 64 concurrent requests to simulate realistic serving conditions, and increase concurrency to 128 for methods with extremely short outputs (e.g., Qwen-DA) to better utilize GPU throughput. Latency is defined as the average end-to-end time from request submission to response completion, while throughput (QPS) is computed as the number of completed requests divided by total wall-clock time. All methods are evaluated under identical hardware and decoding configurations to ensure fair comparison.

\subsection{System-Level Efficiency Analysis}
\label{app:efficient_eval_analysis}
We provide a detailed analysis of system-level efficiency in terms of latency and throughput (QPS) across benchmarks in Table~\ref{tab:main_latency}. While average output token length is a strong proxy for efficiency, we observe that system-level performance is also influenced by the distribution of generation lengths under parallel serving.

Methods such as CoD exhibit unstable behavior on more challenging problems, where occasional long and repetitive reasoning trajectories lead to significant tail latency. Under concurrent execution, these slow requests delay batch completion, resulting in reduced overall throughput despite moderate average token counts. Similarly, ThinkLess enforces strict token limits but often terminates reasoning prematurely or fails to reach valid conclusions, leading to inefficient utilization of the allocated generation budget.

In contrast, DRT produces more compact and well-structured reasoning traces with significantly fewer pathological long outputs. This reduces variance in generation length across samples, leading to more predictable latency and improved batching efficiency. As a result, DRT achieves not only lower average latency but also higher and more stable throughput under concurrent serving.

Overall, these results highlight that improving reasoning efficiency requires not only reducing average token usage, but also controlling the variability of generation length, which plays a critical role in real-world deployment settings.

\section{Robustness to Reward Judge Selection}\label{app:ablation_reward}
\subsection{Ablation Study on Reward Model Size}
To further investigate the impact of reward model size on the effectiveness of our process-level reinforcement learning, we conduct an ablation study by varying the reward model size. Specifically, we evaluate a series of Qwen3 models of different sizes used as reward models, while keeping all other training configurations fixed, allowing us to isolate the effect of reward model size.

\begin{table*}[t]
\centering
\small
\setlength{\tabcolsep}{6pt}
\renewcommand{\arraystretch}{1.15}
\caption{Ablation study on reward model size. We report accuracy (Acc$\uparrow$) on five reasoning benchmarks and their macro average, along with average output tokens (Tokens$\downarrow$).}
\label{tab:ablation_reward_model}
\resizebox{1.0\linewidth}{!}{
\begin{tabular}{lccccccc}
\toprule
\multirow{2}{*}{\textbf{Model Size}} 
& \textbf{MathVista}
& \textbf{MathVerse}
& \textbf{LogicVista}
& \textbf{GSM8K}
& \textbf{Video-Holmes}
& \multicolumn{2}{c}{\textbf{AVG.}} \\
\cmidrule(lr){2-2} \cmidrule(lr){3-3} \cmidrule(lr){4-4} \cmidrule(lr){5-5} \cmidrule(lr){6-6} \cmidrule(lr){7-8}
& Acc$\uparrow$ & Acc$\uparrow$ & Acc$\uparrow$ & Acc$\uparrow$ & Acc$\uparrow$ & Acc$\uparrow$ & Tokens$\downarrow$ \\
\midrule
Qwen3-8B   & 71.9 & 58.6 & 53.2 & 92.1 & 41.6 & 63.5 & 449.0 \\
Qwen3-14B  & 74.6 & 63.0 & 56.8 & 92.3 & 44.2 & 66.2 & 377.5 \\
Qwen3-32B  & 74.9 & 64.1 & 55.7 & 94.2 & 43.3 & 66.5 & 244.3 \\
Qwen3-30B-A3B  & 75.9 & 64.1 & 54.8 & 93.9 & 45.0 & 66.7 & 284.2 \\
Qwen3-235B-A22B & 76.8 & 64.3 & 56.8 & 94.3 & 43.7 & 67.2 & 206.9 \\

\bottomrule
\end{tabular}
}
\vspace{-3mm}
\end{table*}

As shown in Table~\ref{tab:ablation_reward_model}, we observe a clear scaling pattern. Except for the smallest 8B model, all larger reward models (14B and above) achieve performance within 1 point of the strongest 235B model. This suggests that, under the guidance of reference step annotations in DRT-RL, even moderately sized models are sufficient to provide effective process-level supervision, enabling substantial gains in reasoning accuracy.

However, a different trend emerges in terms of efficiency. As the reward model size increases, the average number of output tokens generally decreases, with the 235B model producing the most compact reasoning trajectories. This indicates that while smaller reward models can already capture process-level correctness signals, larger models are more effective at shaping the structure of reasoning.

From the perspective of our reward design, this difference can be attributed to the two components of the reward. The correctness-related signals $R_{\text{main}}$ are relatively easier to learn and can be reliably captured by mid-sized reward models under the guidance of reference step annotations. In contrast, the step bonus $R_{\text{bonus}}$, which encourages depth-aware and efficient exploration, requires finer-grained judgment over the usefulness and necessity of intermediate steps. Such nuanced supervision benefits more from large-scale reward models, which are better equipped to distinguish between redundant, shallow reasoning and compact yet effective multi-step deduction.

Overall, these results highlight a desirable property of our framework: effective process supervision can be achieved with moderate reward model sizes, while scaling up primarily contributes to improving reasoning efficiency and trajectory compactness.

\subsection{Cross-Family Reward Agreement}
We freeze policy rollouts at Steps 0, 100, and 200 of the 200-step DRT-RL process. We use the same randomly sampled 1,000 questions from the RL training set at all three checkpoints, with 16 rollouts per question, and independently rescore them using GPT-5.5 and Gemini-3.5-Flash. The reference traces, judge rubric, reward formula, and rule-based components remain unchanged. We compare the resulting rewards, group-relative advantages, matched-steps MAE and hallucinated-step MAE with those produced by the original Qwen3-235B-A22B judge, as shown in Table~\ref{tab:cross_family_reward_agreement}.

\begin{table}[t]
    \centering
    \caption{
        Cross-family reward and process-judgment agreement on fixed rollouts. All discrepancies and correlations are computed relative to Qwen3-235B-A22B. $\Delta R$ denotes the mean reward difference (judge minus Qwen), and $r$ denotes Pearson correlation.
    }
    \label{tab:cross_family_reward_agreement}
    \small
    \setlength{\tabcolsep}{3pt}
    \renewcommand{\arraystretch}{1.15}
    \resizebox{\linewidth}{!}{%
        \begin{tabular}{ccccccccccc}
            \toprule
            \multirow{2}{*}{\textbf{Judge Model}} & \multirow{2}{*}{\textbf{Step}} & \multicolumn{5}{c}{\textbf{Reward}}
            & \multicolumn{2}{c}{\textbf{Advantage}}
            & \multicolumn{2}{c}{\textbf{Step-count MAE}} \\
            \cmidrule(lr){3-7}
            \cmidrule(lr){8-9}
            \cmidrule(lr){10-11}           
            & & Judge Mean
            & Qwen Mean
            & $\Delta R$
            & MAE
            & $r$
            & MAE
            & $r$
            & Matched
            & Hallucinated \\
            \midrule
            GPT-5.5
            & 0
            & 0.6002
            & 0.6020
            & $-0.0018$
            & 0.1046
            & 0.9477
            & 0.0913
            & 0.9028
            & 1.0638
            & 0.7624 \\

            GPT-5.5
            & 100
            & 0.8043
            & 0.8490
            & $-0.0447$
            & 0.1038
            & 0.9420
            & 0.0796
            & 0.8978
            & 1.2551
            & 0.6640 \\

            GPT-5.5
            & 200
            & 0.8670
            & 0.9204
            & $-0.0534$
            & 0.1044
            & 0.9370
            & 0.0773
            & 0.8933
            & 1.3704
            & 0.5663 \\

            Gemini-3.5-Flash
            & 0
            & 0.6181
            & 0.6020
            & $+0.0161$
            & 0.1063
            & 0.9455
            & 0.0957
            & 0.8940
            & 1.1095
            & 0.7295 \\

            Gemini-3.5-Flash
            & 100
            & 0.8249
            & 0.8490
            & $-0.0241$
            & 0.0995
            & 0.9427
            & 0.0829
            & 0.8914
            & 1.4903
            & 0.6218 \\

            Gemini-3.5-Flash
            & 200
            & 0.8913
            & 0.9204
            & $-0.0291$
            & 0.0977
            & 0.9388
            & 0.0804
            & 0.8844
            & 1.6648
            & 0.5184 \\
            \bottomrule
        \end{tabular}%
    }
    \par\smallskip
\end{table}

Despite minor differences in absolute reward calibration, both judge families remain closely aligned with Qwen throughout training, with reward Pearson correlations of 0.937-0.948 and advantage correlations of 0.884-0.903. Their fine-grained process judgments are also consistent: on trajectories with 7.52 reference steps on average, the matched-step MAE is only 1.06-1.66, while the hallucination-step MAE is 0.52-0.76. Since GRPO is primarily driven by within-group relative advantages, the highly correlated reward and advantage signals indicate that the reward signal is not specific to Qwen and remains robust across judge families as the policy evolves.

To further evaluate end-to-end robustness, we repeat the complete 200-step DRT-RL training using different reward judges while keeping all other configurations unchanged, as shown in Table~\ref{tab:judge_training_robustness}.

\begin{table}[t]
    \centering
    \caption{
        Comparison between different judge models.
    }
    \label{tab:judge_training_robustness}
    \small
    \setlength{\tabcolsep}{3pt}
    \renewcommand{\arraystretch}{1.15}
    \resizebox{\linewidth}{!}{%
        \begin{tabular}{lrrrrrrr}
            \toprule
            \textbf{Reward Judge}
            & \textbf{MathVista}
            & \textbf{MathVerse}
            & \textbf{LogicVista}
            & \textbf{GSM8K}
            & \textbf{Video-Holmes}
            & \textbf{Avg. Acc.$\uparrow$}
            & \textbf{Avg. Tokens$\downarrow$} \\
            \midrule
            Qwen3-30B-A3B
            & 75.9
            & 64.1
            & 54.8
            & 93.9
            & {45.0}
            & 66.7
            & 284.2 \\

            Qwen3-235B-A22B
            & 76.8
            & 64.3
            & 56.8
            & 94.3
            & 43.7
            & 67.2
            & {206.9} \\

            Gemini-3.5-Flash
            & {77.4}
            & 64.3
            & 55.3
            & {94.6}
            & 43.1
            & 66.9
            & 213.2 \\

            GPT-5.5
            & 76.9
            & {65.1}
            & {57.0}
            & {94.6}
            & 43.7
            & {67.5}
            & 259.6 \\
            \bottomrule
        \end{tabular}%
    }
\end{table}

All four judges yield highly consistent final accuracy, with average performance ranging from 66.7 to 67.5. Together with the reward-model-size ablation, these results suggest that once a judge reaches sufficient reasoning capability, it can reliably evaluate intermediate reasoning and provide effective process supervision, making the final policy accuracy largely insensitive to either judge scale or model family. The larger variation in output length indicates that judge selection may influence reasoning compactness more than final accuracy.

\section{Controlled Comparison with Concise CoT}\label{app:concise_cot}
We construct ordinary Concise-CoT baselines using the same source samples, GPT-5.1 generation pipeline, Qwen3-VL-8B initialization, verified RL trajectories, reward judge, and SFT/RL training schedule as DRT. However, during SFT data construction, exact content matching and exact token matching cannot generally be imposed simultaneously: expressing the same atomic reasoning steps as fluent sentence-form CoT requires additional surface tokens. We therefore use two complementary controls and generate two Concise-CoT SFT datasets: (1) token-matched Concise CoT, constrained to the DRT target length and allowed to compress some intermediate verbalization; and (2) content-matched Concise CoT, which preserves the same visual facts, priors, intermediate steps, and answer, but is about 20\% longer than DRT. During RL, both baselines receive only the original questions without DRT-specific prompting and use the same $R_{main}, R_{bonus}$, rollout budget, and optimization settings. We remove only the DRT-specific format reward, with no Concise-CoT-specific reward introduced. The comparison results are shown in Table~\ref{tab:concise_cot_control}.

\begin{table}[t]
    \centering
    \caption{Comparison with Concise-CoT baselines. The token-matched Concise-CoT RL baseline corresponds to VisionR1-dense in Table~\ref{tab:ablation_models}.}
    \label{tab:concise_cot_control}
    \small
    \setlength{\tabcolsep}{3pt}
    \renewcommand{\arraystretch}{1.15}
    \resizebox{\linewidth}{!}{%
        \begin{tabular}{lccccccc}
            \toprule
            \textbf{Model}
            & \textbf{MathVista}
            & \textbf{MathVerse}
            & \textbf{LogicVista}
            & \textbf{Video-Holmes}
            & \textbf{GSM8K}
            & \textbf{Avg. Acc.$\uparrow$}
            & \textbf{Avg. Tokens$\downarrow$} \\
            \midrule
            VisionR1 Raw SFT
            & {71.6}
            & {58.7}
            & {53.2}
            & 40.4
            & {94.8}
            & {63.7}
            & 1233.9 \\

            Concise CoT SFT (token-matched)
            & 71.2
            & 50.9
            & 49.7
            & 39.5
            & 90.8
            & 60.4
            & {150.8} \\

            Concise CoT SFT (content-matched)
            & 70.2
            & 55.8
            & 46.5
            & 38.9
            & 92.2
            & 60.7
            & 199.4 \\

            DRT SFT
            & 71.5
            & 56.2
            & 47.2
            & {42.8}
            & 91.0
            & 61.7
            & 162.6 \\

            \midrule
            VisionR1 Raw RL
            & 72.3
            & 63.4
            & {57.7}
            & 41.4
            & {95.5}
            & 66.1
            & 626.3 \\

            Concise CoT RL (token-matched init.)
            & 73.8
            & 58.5
            & 53.0
            & 42.6
            & 94.8
            & 64.5
            & 253.5 \\

            Concise CoT RL (content-matched init.)
            & 73.0
            & 60.8
            & 52.8
            & 43.5
            & 94.4
            & 64.9
            & 244.2 \\

            DRT RL
            & {76.8}
            & {64.3}
            & 56.8
            & {43.7}
            & 94.3
            & {67.2}
            & {206.9} \\
            \bottomrule
        \end{tabular}%
    }
\end{table}

GPT-5.1 rewriting and generic concision already explain much of the token reduction at the SFT stage, but not the full performance gain: DRT-SFT exceeds the two Concise-CoT controls by 1.0-1.3 average accuracy points. The distinction becomes larger after RL. DRT-RL exceeds the token-matched and content-matched Concise-CoT models by 2.7 and 2.3 points while using 18.4\% and 15.3\% fewer tokens, respectively. It also outperforms the long-CoT RL baseline by 1.1 points with approximately 67\% fewer tokens. The gains over both Concise-CoT controls are consistent across all four multimodal benchmarks, while performance on the text-only GSM8K benchmark remains comparable. This pattern supports the intended role of DRT: separating visual evidence from deduction reduces their conflation, while compact, addressable transitions better align with step matching, hallucination correction, and effective-step rewards. Thus, GPT-based concise rewriting explains part of the compression benefit, but the controlled results show that DRT contributes beyond generic concision and yields a stronger accuracy-efficiency tradeoff.

\section{Limitations}\label{app:limitation}
Despite the strong performance achieved by DRT in improving reasoning efficiency, several avenues remain for future exploration. The current DRT formulation mainly follows a sequential chain-style reasoning process. Although such a design improves reasoning density and inference efficiency, it does not explicitly support more flexible graph-structured reasoning patterns involving branching exploration or dependency-aware reasoning. Extending DRT toward more adaptive and structured reasoning paradigms remains an important future direction.

\section{Societal Impacts}\label{app:Societal_Impacts}
1) This work studies efficient multimodal reasoning through Dense Reasoning Trace (DRT), which aims to reduce the computational cost of explicit reasoning while maintaining reasoning accuracy and visual grounding. By improving reasoning efficiency, our approach may help lower inference latency and resource consumption for large multimodal models, potentially making advanced reasoning systems more accessible and environmentally sustainable.
2) Similar to other MLLMs, DRT-based models may be vulnerable to adversarial prompting that results in harmful or inappropriate outputs. This highlights the ongoing need for advances in AI safety to ensure responsible and secure use of such models.

\section{AI Use Statement}
We used large language models to assist with language polishing. We also used GPT-5.1 to generate and verify synthetic reasoning traces, and LLM-based judges for reward computation and step-level evaluation, as described in the methodology and appendices. The authors reviewed and revised the AI-assisted text and take responsibility for the final manuscript, methods, results, and research artifacts.

\section{Licenses of Existing Assets}\label{app:license}


\begin{table}[h]
\centering
\small
\setlength{\tabcolsep}{5pt}
\renewcommand{\arraystretch}{1.1}
\caption{Licenses of primary existing assets used in this work.}
\resizebox{1.0\linewidth}{!}{
\begin{tabular}{l l l}
\toprule
Asset & License & URL \\
\midrule

Qwen3-VL~\cite{bai2025qwen3} & Apache License 2.0 &
\url{https://huggingface.co/collections/Qwen/qwen3-vl} \\

vLLM~\cite{kwon2023efficient} & Apache License 2.0 &
\url{https://github.com/vllm-project/vllm} \\

ms-swift~\cite{ms-swift} & Apache License 2.0 &
\url{https://github.com/modelscope/ms-swift} \\

verl~\cite{sheng2024hybridflow} & Apache License 2.0 &
\url{https://github.com/volcengine/verl} \\

DAPO~\cite{yu2025dapo} & Apache License 2.0 &
\url{https://huggingface.co/datasets/BytedTsinghua-SIA/DAPO-Math-17k} \\

Vision-R1-cold~\cite{huang2025vision} & cc-by-nc-sa-4.0 &
\url{https://huggingface.co/datasets/Osilly/Vision-R1-cold} \\

Vision-R1-rl~\cite{huang2025vision} & cc-by-nc-sa-4.0 &
\url{https://huggingface.co/datasets/Osilly/Vision-R1-rl} \\

\bottomrule
\end{tabular}
}
\label{tab:licenses}
\end{table}

\section{Details of Dataset Generation Prompts}\label{app:dataset_prompt}

\subsection{Prompts for DRT-SFT Dataset Construction}\label{app:sft_dataset_prompt}

\begin{promptbox}[DRT-SFT Trajectory Generation Prompt]
You are an expert in optimizing Chain-of-Thought (CoT) efficiency for VLM.
Your task is to compress the Assistant's response into a structured, high-density format.

--- Input Data ---
[Question]: 
{question_content}

[Original Response]:
{original_response}

--- Task Requirements ---
1. **Deconstruct**: Analyze the [Original Response] to separate visual observations, logical reasoning, and the final answer.
2. **Compress**: Rewrite the content into a "Dense Cognitive Trace".
   - **Style**: Telegraphic (arrows ->, symbols, short phrases). NO fillers.
   - **Visual**: Extract visual evidence mentioned in the thought process.
   - **Think**: 
     - **Step 1 (Priors)**: At the very beginning, briefly list necessary commonsense or formulas (e.g., "Area=pi*r^2", "Boiling point=100 deg C") if applicable.
     - **Step 2 (Logic)**: Compress the subsequent logical steps.
3. **Format**: The "dense_version" string MUST strictly follow this XML structure:
   <visual>...concise visual evidence...</visual><think>...[Priors] -> ...compressed reasoning...</think><answer>...final answer...</answer>

--- Output Format ---
Return a strictly valid JSON object:
{{
    "original_length": <int, char count of original text>,
    "dense_version": "<string, the compressed text containing <visual>, <think>, and <answer> tags>",
    "dense_length": <int, char count of dense version>,
    "efficiency_ratio": <float, dense_length / original_length>,
    "waste_analysis": "<string, brief comment on what was removed>"
}}
\end{promptbox}

\newpage
\subsection{Prompts for DRT-RL Dataset Construction}\label{app:rl_dataset_prompt}

\begin{promptbox}[DRT-RL Trajectory Generation Prompt for Multimodal Samples]
Solve the following visual math problem step-by-step.

Problem: {question}

Requirements:
1. Use only evidence visible in the provided image(s) and text.
2. Generate a concise but logically complete reasoning trace.
3. If a detail is not clearly supported by the image/text, omit it instead of guessing.
4. Prefer fewer grounded steps over speculative steps.
5. Output strictly in JSON format:
{{
  "steps": ["Step 1...", "Step 2..."],
  "final_answer": "The final answer"
}}
\end{promptbox}

\begin{promptbox}[DRT-RL Trajectory Generation Prompt for Text-only Samples]
Solve the following math problem step-by-step.

Problem: {question}

Requirements:
1. Use only evidence from the problem statement and valid math reasoning.
2. Generate a concise but logically complete reasoning trace.
3. Output strictly in JSON format:
{{
  "steps": ["Step 1...", "Step 2..."],
  "final_answer": "The final answer"
}}
\end{promptbox}

\begin{promptbox}[DRT-RL Trajectory Verification Prompt]
Question: {question}
Correct Answer: {answer}
Generated Answer: {pred_answer}
Candidate Trajectory: {json.dumps(steps, ensure_ascii=False)}

You are auditing a candidate stepwise reasoning trajectory conditioned on the question and the ground-truth answer. A correct final answer alone is not sufficient. Accept the trajectory only if it is reliable across perception, intermediate reasoning, and final answer.

Audit dimensions:
1. `visual_fidelity`: For image-based problems, every perceptual claim must be directly supported by explicit visual/textual evidence, such as visible labels, symbols, quantities, spatial relations, or clearly specified attributes. Mark false if the trajectory invents or assumes visual values, diagram properties, counts, relations, or measurements that are not observable. For text-only problems, mark true unless the trajectory makes unsupported visual/perceptual claims.
2. `reasoning_faithfulness`: Each deductive step must follow from verified observations, the problem statement, explicitly introduced priors, or established theorems. Mark false for logical inconsistencies, circular reasoning, speculative shortcuts, ad hoc assumptions, unsupported intermediate values, use of the ground-truth answer as a premise, or traces that merely restate the answer without a non-trivial derivation.
3. `semantic_answer_correctness`: Mark true if the Generated Answer is mathematically or semantically equivalent to the Correct Answer, allowing equivalent expressions, units, or formatting.

Decision rule: `verdict` must be `PASS` only when all three audit dimensions are true; otherwise it must be `FAIL`.

Output strict JSON only:
{{
  "visual_fidelity": true,
  "reasoning_faithfulness": true,
  "semantic_answer_correctness": true,
  "verdict": "PASS",
  "reason": "brief explanation"
}}
\end{promptbox}

\section{VERL Judge Prompt}\label{app:verl_judge_prompt}

\begin{promptbox}[LLM Judge Prompt]
You are a strict Math Teacher grading a student's reasoning for a {problem_type}.

**Original Problem:**
{problem_text}

**Reference Solution (Standard Steps):**
{gt_steps_str}

**Student's Thinking Process:**
{pred_think}

**Student's Final Answer:**
{pred_final}

**Grading Tasks:**

1. **Calculate matched_step_count:**
A reference step is counted as matched if the student correctly reproduces the core logical transition or intermediate result, even if expressed differently or with different granularity (e.g., merged or split steps).
Matched steps are counted even if hallucination exists; hallucination is tracked separately.

2. **Calculate effective_reasoning_step_count:**
Count distinct, non-trivial reasoning steps that contribute to reaching the final solution (not merely intermediate support).
Do NOT count repetition, filler, paraphrases, or artificial over-splitting.

3. **Calculate deep_exploration_step_count:**
Count reasoning steps that are NOT strictly required to reproduce the reference solution but remain logically valid and relevant to the problem (e.g., verification, alternative reasoning, elimination of alternatives).
All deep exploration steps MUST also be included in effective_reasoning_step_count.

4. **Do NOT estimate target counts:**
The number of reference steps serves as the target for effective reasoning, while deep exploration is externally capped during reward computation.

5. **Determine hallucination within matched steps only:**
A hallucinated matched step is one that relies on unsupported assumptions, invented values, ungrounded visual claims, or unjustified logical inferences.
Report both hallucinated_matched_step_count and has_hallucination_in_matched_steps.

6. **Counting rules:**
- All counts must be non-negative integers.
- Do not double count the same reasoning step across categories (except that deep exploration is a subset of effective steps).

7. **Output JSON ONLY with the following keys:**
matched_step_count,
total_reference_steps,
effective_reasoning_step_count,
deep_exploration_step_count,
hallucinated_matched_step_count,
has_hallucination_in_matched_steps,
reason

The reason field must be concise (1--3 sentences) explaining key grading decisions.
\end{promptbox}

\section{Evaluation prompts}~\label{app:evaluation_prompts}

\begin{promptbox}[Qwen3-VL Standard Prompt]
{question}
\end{promptbox}

\begin{promptbox}[Direct-Answer]
{question}

Please answer concisely with short words or phrases when possible.
\end{promptbox}

\begin{promptbox}[Dense Reasoning Trace]
{question}

Analyze this question to provide a **Dense Cognitive Trace**.

--- Requirements ---
1. **Format**: Use a **telegraphic style** (concise phrases, arrows '->', symbols). NO conversational fillers.
2. **Structure**: Your response MUST strictly follow this XML structure:
   <visual>...concise visual evidence...</visual><think>...[Priors] -> ...compressed reasoning...</think><answer>...final answer...</answer>
3. **Content**: Deconstruct into visual observations, logical reasoning, and the final answer.
\end{promptbox}

\section{Step-level GPT Analysis}\label{app:step-level-gpt-judge}
\begin{promptbox}[Reason Step Count]
You are a strict reasoning annotator for a visual math benchmark.

Task: count the student's `reason_step_count`.

Definition:
- `reason_step_count` means the number of distinct atomic reasoning acts in the student's response.
- Count semantically distinct atomic acts such as:
  1. reading or restating a given condition if it is used in the reasoning flow,
  2. extracting a visual fact,
  3. making an assumption,
  4. making a guess or hypothesis,
  5. deriving an intermediate relation,
  6. checking or confirming a previous result,
  7. revising a previous inference after recalculation.
- In DRT-style traces, an atomic act often looks like one `->` segment, but do NOT rely only on formatting.
- In long natural-language responses, judge semantically.
- Do NOT count pure filler, repeated paraphrases of the exact same act, stylistic padding, or copied boilerplate.

Original Problem: {problem_text}

Student Reasoning: {pred_think}

Student Final Answer: {pred_final}

Output strict JSON only:
{{
  "reason_step_count": 0,
  "reason": "brief explanation"
}}
\end{promptbox}

\begin{promptbox}[Effective Reason Step Count]
You are a strict process-reward annotator for a visual math benchmark.

Task: count the student's `effective_reasoning_step_count`.

Definition:
- `effective_reasoning_step_count` is narrower than raw reasoning-step count.
- A step counts as effective only if it is a distinct, meaningful, solution-advancing process step.
- Count a step as effective if either:
  1. it matches a useful intermediate result / logical milestone in the GT reference steps, EVEN IF the step's derivation contains hallucination, unsupported assumption, or guessing, OR
  2. it belongs to a different but still meaningful reasoning path that genuinely supports deriving the answer.
- This metric is intended to approximate "process reward".
- Strictly do NOT give credit to water processes:
  - filler,
  - repetition,
  - circular restatement,
  - verbose rephrasing without new progress,
  - dead-end wandering that does not help derive the answer.

Original Problem: {problem_text}

Reference GT Steps: {gt_steps_str}

Student Reasoning: {pred_think}

Student Final Answer: {pred_final}

Output strict JSON only:
{{
  "effective_reasoning_step_count": 0,
  "reason": "brief explanation"
}}
\end{promptbox}

\begin{promptbox}[Hallucination Effective Reason Step Count]
You are a strict hallucination annotator for a visual math benchmark.

Task: count the student's `hallucinated_effective_step_count`.

Definition:
- Only consider steps that would qualify as effective reasoning steps under this definition:
  distinct, meaningful, solution-advancing process steps that either match GT intermediate milestones
  or form a genuinely useful alternative path toward the answer.
- Among those effective steps only, count how many contain hallucination-like behavior:
  1. unsupported guesses,
  2. unjustified assumptions,
  3. fabricated visual observations,
  4. invented numeric values,
  5. logical leaps not rigorously derived from the problem.
- Do NOT count hallucinations that occur only inside non-effective filler text.
- The count must be based on the effective-step subset, not the whole response.

Original Problem: {problem_text}

Reference GT Steps: {gt_steps_str}

Student Reasoning: {pred_think}

Student Final Answer: {pred_final}

Output strict JSON only:
{{
  "hallucinated_effective_step_count": 0,
  "reason": "brief explanation"
}}
\end{promptbox}

\newpage
\section*{NeurIPS Paper Checklist}

\begin{enumerate}

\item {\bf Claims}
    \item[] Question: Do the main claims made in the abstract and introduction accurately reflect the paper's contributions and scope?
    \item[] Answer: \answerYes{} 
    \item[] Justification: The abstract and introduction accurately summarize the main contributions, scope, and empirical findings of the paper.

\item {\bf Limitations}
    \item[] Question: Does the paper discuss the limitations of the work performed by the authors?
    \item[] Answer: \answerYes{} 
    \item[] Justification: The paper discusses the limitations of the work in Appendix~\ref{app:limitation}.

\item {\bf Theory assumptions and proofs}
    \item[] Question: For each theoretical result, does the paper provide the full set of assumptions and a complete (and correct) proof?
    \item[] Answer: \answerNA{} 
    \item[] Justification: The paper does not include theoretical results, such as theorems or formal proofs, as it focuses on methodological contributions and empirical validation.

    \item {\bf Experimental result reproducibility}
    \item[] Question: Does the paper fully disclose all the information needed to reproduce the main experimental results of the paper to the extent that it affects the main claims and/or conclusions of the paper (regardless of whether the code and data are provided or not)?
    \item[] Answer: \answerYes{} 
    \item[] Justification: The paper provides all necessary details to reproduce the main experimental results, including method descriptions (Section~\ref{method}), prompts (Appendix~\ref{app:dataset_prompt},~\ref{app:verl_judge_prompt},~\ref{app:evaluation_prompts},~\ref{app:step-level-gpt-judge}), and training setting (Appendix~\ref{app:verl_hyperparameters},~\ref{app:training}).

\item {\bf Open access to data and code}
    \item[] Question: Does the paper provide open access to the data and code, with sufficient instructions to faithfully reproduce the main experimental results, as described in supplemental material?
    \item[] Answer: \answerNo{} 
    \item[] Justification: We will release the training and evaluation code, configuration files, prompts, and inference scripts upon acceptance.

\item {\bf Experimental setting/details}
    \item[] Question: Does the paper specify all the training and test details (e.g., data splits, hyperparameters, how they were chosen, type of optimizer) necessary to understand the results?
    \item[] Answer: \answerYes{} 
    \item[] Justification: We provide training and evaluation prompts in Appendix~\ref{app:verl_judge_prompt},~\ref{app:evaluation_prompts},~\ref{app:step-level-gpt-judge} and training setting in Appendix~\ref{app:verl_hyperparameters},~\ref{app:training}.

\item {\bf Experiment statistical significance}
    \item[] Question: Does the paper report error bars suitably and correctly defined or other appropriate information about the statistical significance of the experiments?
    \item[] Answer: \answerNo{} 
    \item[] Justification: Due to the limited computational resources, we do not conduct sufficient independent runs to report error bars for training and evaluation.

\item {\bf Experiments compute resources}
    \item[] Question: For each experiment, does the paper provide sufficient information on the computer resources (type of compute workers, memory, time of execution) needed to reproduce the experiments?
    \item[] Answer: \answerYes{} 
    \item[] Justification: We provide sufficient information on the computer resources in Appendix~\ref{app:training} and Appendix~\ref{app:efficient_eval_setting}.
    
\item {\bf Code of ethics}
    \item[] Question: Does the research conducted in the paper conform, in every respect, with the NeurIPS Code of Ethics \url{https://neurips.cc/public/EthicsGuidelines}?
    \item[] Answer: \answerYes{} 
    \item[] Justification: We have reviewed the NeurIPS Code of Ethics and comply with it.

\item {\bf Broader impacts}
    \item[] Question: Does the paper discuss both potential positive societal impacts and negative societal impacts of the work performed?
    \item[] Answer: \answerYes{} 
    \item[] Justification: We report societal impacts in \cref{app:Societal_Impacts}.
    
\item {\bf Safeguards}
    \item[] Question: Does the paper describe safeguards that have been put in place for responsible release of data or models that have a high risk for misuse (e.g., pre-trained language models, image generators, or scraped datasets)?
    \item[] Answer: \answerNA{} 
    \item[] Justification: This work focuses on improving the efficiency and grounding of multimodal reasoning models through structured reasoning representations and reinforcement learning. It does not introduce a new foundation model, internet-scale scraped dataset, or generative system that poses significant risks of misuse beyond those already associated with the underlying open-source models.

\item {\bf Licenses for existing assets}
    \item[] Question: Are the creators or original owners of assets (e.g., code, data, models), used in the paper, properly credited and are the license and terms of use explicitly mentioned and properly respected?
    \item[] Answer: \answerYes{} 
    \item[] Justification: We provide the license in Appendix~\ref{app:license}.

\item {\bf New assets}
    \item[] Question: Are new assets introduced in the paper well documented and is the documentation provided alongside the assets?
    \item[] Answer: \answerYes{} 
    \item[] Justification: The paper introduces new assets, including two datasets and a multimodal large language model. We will provide comprehensive documentation alongside the release, including information about data sources, licensing and usage guidelines when open-sourcing the assets.

\item {\bf Crowdsourcing and research with human subjects}
    \item[] Question: For crowdsourcing experiments and research with human subjects, does the paper include the full text of instructions given to participants and screenshots, if applicable, as well as details about compensation (if any)? 
    \item[] Answer: \answerNA{} 
    \item[] Justification: This work does not involve crowdsourcing experiments or research with human subjects.

\item {\bf Institutional review board (IRB) approvals or equivalent for research with human subjects}
    \item[] Question: Does the paper describe potential risks incurred by study participants, whether such risks were disclosed to the subjects, and whether Institutional Review Board (IRB) approvals (or an equivalent approval/review based on the requirements of your country or institution) were obtained?
    \item[] Answer: \answerNA{} 
    \item[] Justification: This work does not involve crowdsourcing experiments or research with human subjects, and therefore IRB approval was not required.

\item {\bf Declaration of LLM usage}
    \item[] Question: Does the paper describe the usage of LLMs if it is an important, original, or non-standard component of the core methods in this research? Note that if the LLM is used only for writing, editing, or formatting purposes and does \emph{not} impact the core methodology, scientific rigor, or originality of the research, declaration is not required.
    \item[] Answer: \answerYes{} 
    \item[] Justification: This work uses large language models as important components of the proposed methodology. Specifically, GPT-5.1 is used to generate and verify stepwise reasoning trajectories during the construction of the DRT-SFT and DRT-RL dataset, and Qwen3-235B is used as a step-level judge for reward computation during reinforcement learning. These usages are described in the main paper and supplemental material.

\end{enumerate}

\end{document}